\documentclass[sigplan,10pt,authorversion]{acmart}
\renewcommand\footnotetextcopyrightpermission[1]{%
  \footnotetext{Preprint. Accepted to the 40th ACM International Conference on Supercomputing (ICS 2026).}%
}
\AtBeginDocument{%
  }

\usepackage{enumitem}

\usepackage[endLComment=]{algpseudocodex}
\usepackage[ruled]{algorithm}

\usepackage{amsmath}
\allowdisplaybreaks
\usepackage{subcaption} 
\usepackage{threeparttable}
\usepackage{multirow}
\usepackage{booktabs}
\usepackage{enumitem}
\usepackage{xcolor}
\usepackage{bm}

\usepackage{amsthm,thmtools}
\declaretheoremstyle[
  headfont=\normalfont\itshape\bfseries,
  headpunct=\textup{:},
  bodyfont=\normalfont,
]{myremark}

\declaretheoremstyle[
  headfont=\normalfont\itshape,
  headpunct=:,
  bodyfont=\normalfont,
]{myremarknonum}

\theoremstyle{myremark}
\newtheorem{remark}{Remark}
\theoremstyle{myremarknonum}
\newtheorem*{remark*}{Remark}

\usepackage[compact]{titlesec}
\titlespacing{\section}{0pt}{2ex}{1ex}
\titlespacing{\subsection}{0pt}{2ex}{1ex}
\titlespacing{\subsubsection}{0pt}{0.5ex}{1ex}

\begin{document}

\title{\emph{Rudder}: Steering Prefetching in Distributed \\ GNN Training using LLM Agents} 

\author{Aishwarya Sarkar}
\affiliation{%
  \institution{Iowa State University}
  \city{Ames}
  \state{IA}
  \country{USA}
}
\email{asarkar1@iastate.edu}

\author{Sayan Ghosh}
\affiliation{%
  \institution{Pacific Northwest National Laboratory}
  \city{Richland}
  \state{WA}
  \country{USA}
}
\email{sayan.ghosh@pnnl.gov}

\author{Nathan Tallent}
\affiliation{%
  \institution{Pacific Northwest National Laboratory}
  \city{Richland}
  \state{WA}
  \country{USA}
}
\email{tallent@pnnl.gov}

\author{Aman Chadha}
\affiliation{%
  \institution{Amazon GenAI}
  \city{Cupertino}
  \state{CA}
  \country{USA}
}
\email{hi@aman.ai}

\author{Tanya Roosta}
\affiliation{%
  \institution{University of California, Berkeley}
  \city{Berkeley}
  \state{CA}
  \country{USA}
}
\email{troosta@ischool.berkeley.edu}

\author{Ali Jannesari}
\affiliation{%
  \institution{Iowa State University}
  \city{Ames}
  \state{IA}
  \country{USA}
}
\email{jannesari@iastate.edu}




\newcommand{\note}[1]{\textcolor{red}{\textbf{[}#1\textbf{]}}}

\newcommand{\nrt}[1]{\textcolor{blue}{\textbf{nrt:}#1\textbf{}}}

\begin{abstract}
Large-scale Graph Neural Networks (GNNs) are typically trained by sampling a vertex's neighbors to a fixed distance.
Because large input graphs are distributed, training requires frequent irregular communication that stalls forward progress.
Moreover, fetched data changes with graph, graph distribution, sample and batch parameters, and caching polices.
Consequently, \emph{any} static prefetching method will miss crucial opportunities to adapt to different dynamic conditions.
In this paper, we introduce \emph{Rudder}, a software module embedded in the state-of-the-art AWS DistDGL framework, to autonomously prefetch remote nodes and minimize communication.
Rudder's adaptation contrasts with both standard heuristics and traditional ML classifiers.
We observe that the generative AI found in contemporary Large Language Models (LLMs) exhibits emergent properties like In-Context Learning (ICL) for zero-shot tasks, with logical multi-step reasoning.
We find this behavior well-suited for adaptive control \emph{even with substantial undertraining}.
Evaluations using standard datasets and \emph{unseen configurations} on the NERSC Perlmutter supercomputer show up to 91\% improvement in end-to-end training performance over baseline DistDGL (no prefetching), and an 82\% improvement over static prefetching, reducing communication by over 50\%. Our code is available at \href{https://github.com/aishwaryyasarkar/rudder-llm-agent}{github.com/aishwaryyasarkar/rudder-llm-agent}.

\end{abstract}
\maketitle
\thispagestyle{plain}
\pagestyle{plain}

\section{Introduction}
Graph Neural Networks (GNNs) are essential for learning with unstructured data across applications from recommendation systems, scientific simulations, to life sciences~\cite{zhou2020graph, wu2020comprehensive, li2021dgl, jiang2022graph}.
Real-world graphs often grow large~\cite{leskovec2007graph} and require distributed memory, requiring partitioning the input graphs across several Processing Elements (PEs)~\cite{shao2024distributed}.
Moreover, as a GNN processes an input graph, it requires knowledge of each vertex's neighborhood---all neighbors to a given distance---meaning that a single vertex may require communication of neighbor vertices from \emph{any or all} other partitions.
Consequently, unstructured real-world graphs cause unpredictable and unbounded communication.
To avoid neighborhood explosion~\cite{zhou2020graph}, a random ``sample'' of the distance-\emph{k} neighborhood is used instead.
However, even with sampling, the size and constituents of the neighborhood change, resulting in load imbalance caused by (a) varying communication delays and (b) neighborhood set sizes.

\begin{figure}[!ht]
  \centering
  \includegraphics[width=\linewidth]{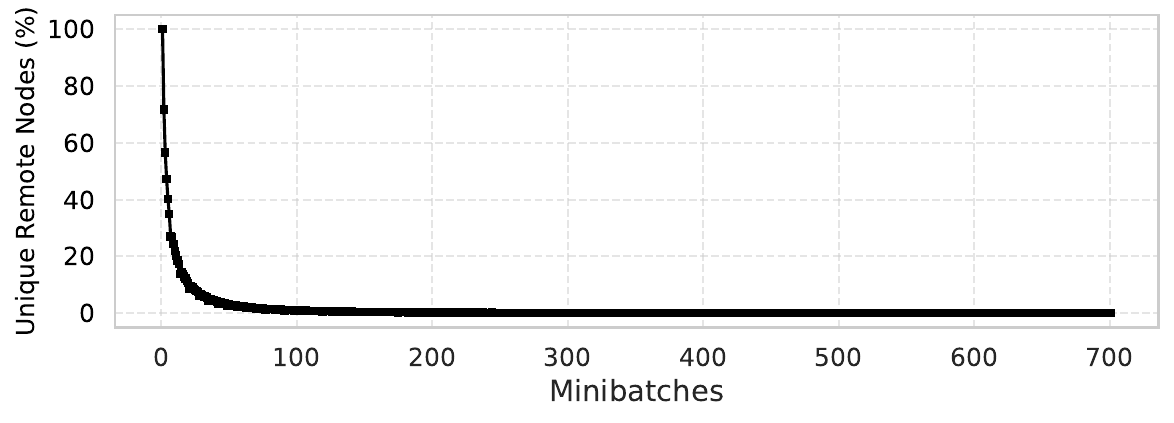}
  \caption{Declining unique remote nodes in GNN training.}
 \label{fig:prefetching-motivation}
\end{figure}

To avoid some communication delays, recent work has explored forms of prefetching, or caching the remote nodes (without affecting accuracy)~\cite{park2024lsm,sarkar2024massivegnn,song2023granndis,kaler2023communication,lin2020pagraph,yang2022gnnlab,liu2023bgl,zhu2019aligraph}.
Fig.~\ref{fig:prefetching-motivation} shows that the amount of sampled unique remote nodes decreases as minibatches progress, showing the potential for effective prefetching schemes.

Due to neighborhood explosion, prefetching everything at once is unfeasible (memory constraints, latency).
A practical solution maintains a fixed-size local persistent buffer 
and occasionally replaces vertices that will be unused.
The challenge that any effective prefetching scheme must address is the compound question of \emph{what and when to replace}, which requires solving a dynamic optimization problem consisting of varying
\emph{graph, partitioning, neighbor sampling, batch size, and prefetching parameters}.

Recent works~\cite{park2024lsm,sarkar2024massivegnn,song2023granndis,kaler2023communication} have proposed prefetching frequently accessed features of remote nodes (asynchronously pipelining associated data transfers) at different levels to mitigate the communication overheads (considering both network and CPU--GPU data transfers) for GNN training, without focusing on \emph{replacement}. Existing static prefetching policies~\cite{sarkar2024massivegnn, kaler2023communication} require costly trial-and-error to pre-process and find optimal parameters given fixed assumptions (i.e., partitioning, architecture hyperparameters, etc.) for every dataset and training configurations. 

Machine learning (ML) approaches improved data persistence and locality for Content Delivery Networks (CDNs)~\cite{liu2017content, narayanan2018deepcache, song2023halp}. However, none of the existing approaches  apply to GNN prefetching. 
Moreover, the dynamic nature of distributed GNN training highlights the fundamental challenge of \emph{collecting ground truth training data, e.g., classifying replacement outcomes over the range of configurations.}

Generative AI has demonstrated remarkable achievements in a wide range of tasks, including text generation, summarization, reasoning, etc~\cite{chang2024survey}. Recent advances and wide availability of LLMs have shifted attention to autonomous \emph{AI agents}, integrating them as reasoning components into the system that can interface with external tools or systems, observe their states, making autonomous decisions without task-specific training. 
Such agents can use internal knowledge of the pre-trained LLM to adapt to unseen tasks through In-Context Learning (ICL)~\cite{brown2020language}, enabling \emph{zero-shot learning}, where the model generalizes to unseen tasks based solely on instructions or structured prompts. 

\begin{figure}[t]
  \centering
  \includegraphics[width=\linewidth]{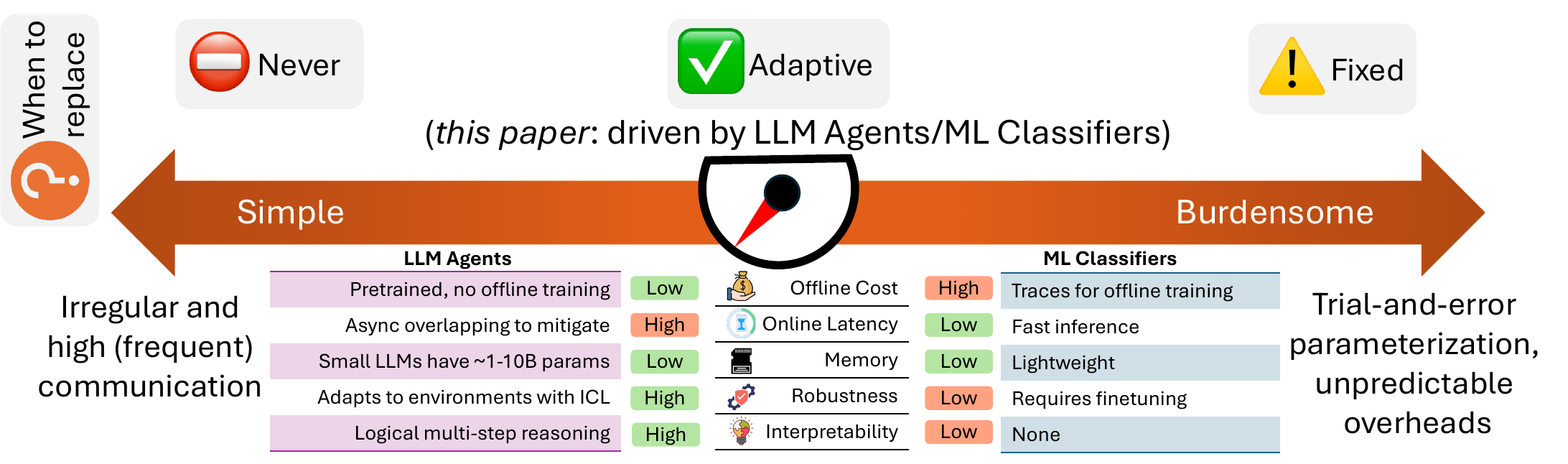}
\caption{Prefetching interfaces can range from \emph{simple} to \emph{burdensome}. 
The simplest designs sacrifice performance; while the most burdensome require enormous tuning.
Rudder achieves high performance while requiring little tuning.
}
\label{fig:scenarios}
\end{figure}

In this paper, we design and deploy an \emph{adaptive replacement strategy}, namely \emph{Rudder}, with real-time data persistence steering to optimize communication for distributed sampling-based GNN workloads, within the popular AWS DistDGL distributed GNN framework. 
Rudder has been carefully designed to achieve superior adaptive prefetching performance 
while (a) avoiding burdensome tuning and configuration
and (b) imposing modest and hidden computational overhead.
Fig.~\ref{fig:scenarios} depicts trade-offs with prefetching strategies and summarizes Rudder's design choices. 

In summary, our contributions are as follows.
\begin{itemize}[leftmargin=*,nosep]
\item 
Design adaptive prefetching, based on in-context learning, to improve load balance by reducing exposed network communication. 
\emph{Rudder}'s prefetching overlaps with model training and is usually fully hidden, improving end-to-end GNN training execution times up to 90\%.

\item Extensive design study (\S\ref{ssec:motiv-suitability}, \ref{sec:method-design}) that compares adaptive prefetching based on (a) in-context learning with LLMs vs. (b) several ML classifiers designed specifically for \emph{Rudder}.
Both sets of designs use real-time performance metrics. 





\item Extensive evaluation using diverse graph datasets on the NERSC Perlmutter platform that includes: performance\slash quality trade-offs, scalability analysis, suitability of latest LLMs as agents, failure modes, out-of-distribution studies, and comparisons with ML classifiers (\S\ref{sec:results}).
\end{itemize}

\section{Background and Motivation}
\label{sec:motive}

We examine the key questions of prefetching for GNN workloads. 
First we consider \emph{what}-to-replace (\S\ref{ssec:motiv-replacement}).
Next in \S\ref{ssec:motiv-suitability}, we posit dynamically determining \emph{when}-to-replace. 
\subsection{Prefetching and Replacement Strategies}
\label{ssec:motiv-replacement}

\begin{figure}[!ht]
  \centering
  \includegraphics[width=\linewidth]{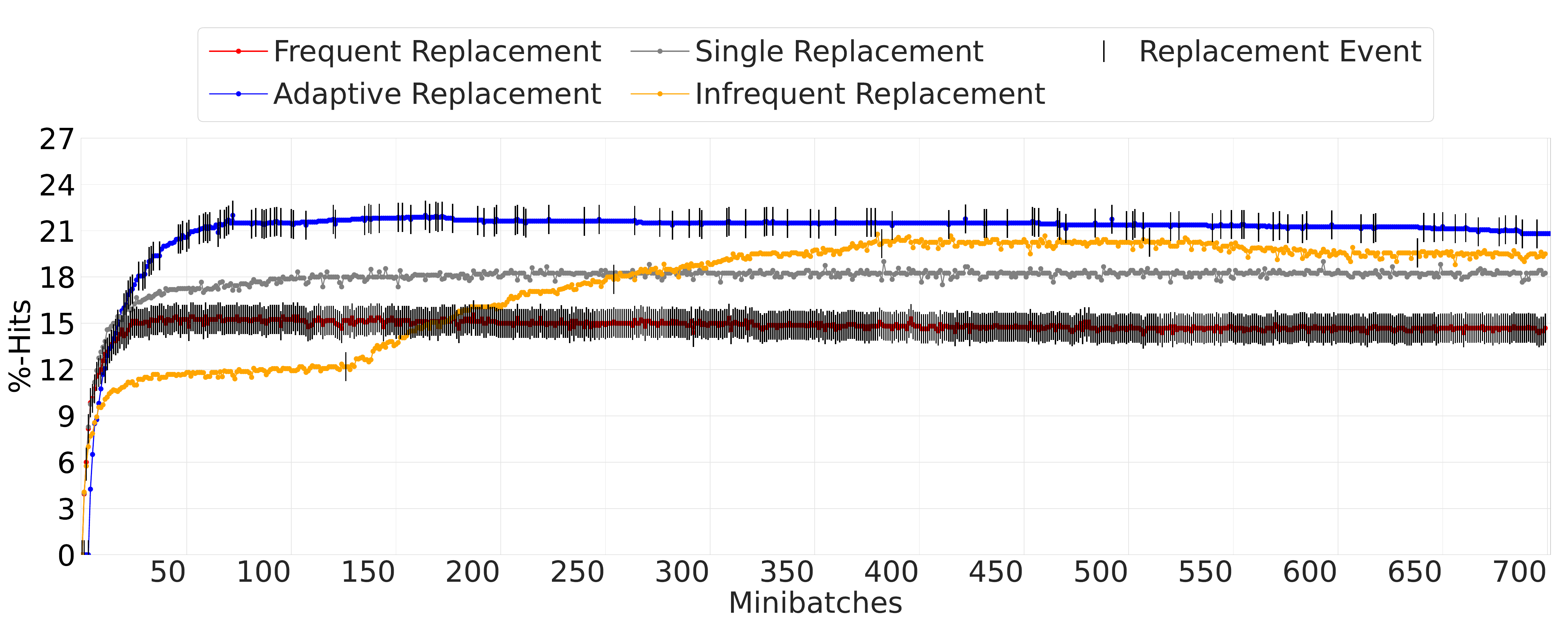}
  \caption{Adaptive replacement consistently yields best \%-Hits (\emph{higher} is better), relative to other replacement strategies.}
 \label{fig:node-hits}
\end{figure}

Although several forms of prefetching to optimize data transfers are prevalent in contemporary GNN frameworks~\cite{sarkar2024massivegnn, graphscale, zheng2020distdgl, Kaler2023.SALIENT++}), an \emph{adaptive replacement strategy} to improve the overall data persistence (thereby minimizing communication) throughout diverse sampling patterns currently does not exist. Strategies based on fixed policies or heuristics lack the ability to quickly adapt to different inputs, node configurations and execution performance. Checking for replacements at every minibatch is a reasonable middle-ground when starting with an empty buffer (especially during the expensive initial stages), compared to single and infrequent replacements---both increase communication and are negatively impacted by the staleness of data, as demonstrated in Fig.~\ref{fig:node-hits} via the ``\%-Hits'' metric (higher is better) on 4 nodes of NERSC Perlmutter, calculated as percentage of remote nodes in the local persistent buffer. Consequently, we observe better data persistence with the proposed adaptive replacement relative to existing approaches.

\begin{figure}[!ht]
  \centering
  \includegraphics[width=0.8\linewidth]{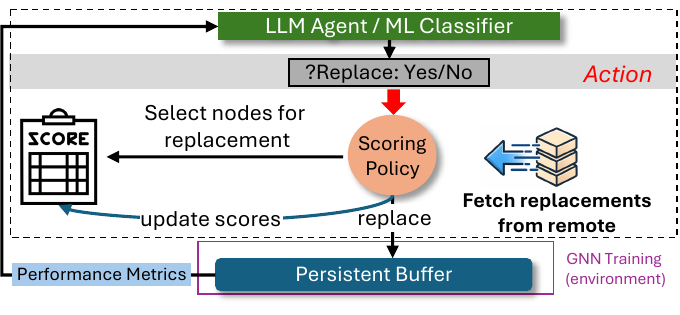}
\caption{High-level aspects of our replacement strategy based on a scoring policy which tracks recent usage.}
\label{fig:scoring}
\end{figure}

Periodically, the ``stale'' nodes (i.e., unused nodes during recent training epochs) are replaced to make room for more ``relevant'' ones in the limited persistent buffer. Our mechanism for identifying prospective nodes for replacement is based on frequency tracking, but is more aggressive than existing caching policies such as Least Frequently Used (LFU). In LFU variants, cache pollution is likely (items with short-lived popularity exhibit higher frequency counts and longer persistence), whereas our policy penalizes stasis to frequently refresh the persistent buffer (see Fig.~\ref{fig:scoring}). 
When an item is accessed, associated frequency score is incremented by $1$. Conversely, we penalize items (multiplying scores by $0.95$) that are not accessed during the current minibatch-sampling epoch. Items whose scores fall below $0.95$ (i.e., ``stale'' nodes) are replaced with recently sampled remote nodes (if no ``stale'' nodes, replacement is skipped). 
\subsection{Intelligent Prefetching Controllers}
\label{ssec:motiv-suitability}
Our adaptive replacement strategy is derived from latest and traditional ML approaches: LLM agents (stateful generative models inside an agentic loop) and ML Classifiers (stateless discriminative models that map current buffer statistics to a binary decision). We discuss key considerations.
\subsubsection{LLM agents vs. ML classifiers}\label{sssec:agents_classifiers}
\begin{figure}[!ht]
  \centering
  \includegraphics[width=\linewidth]{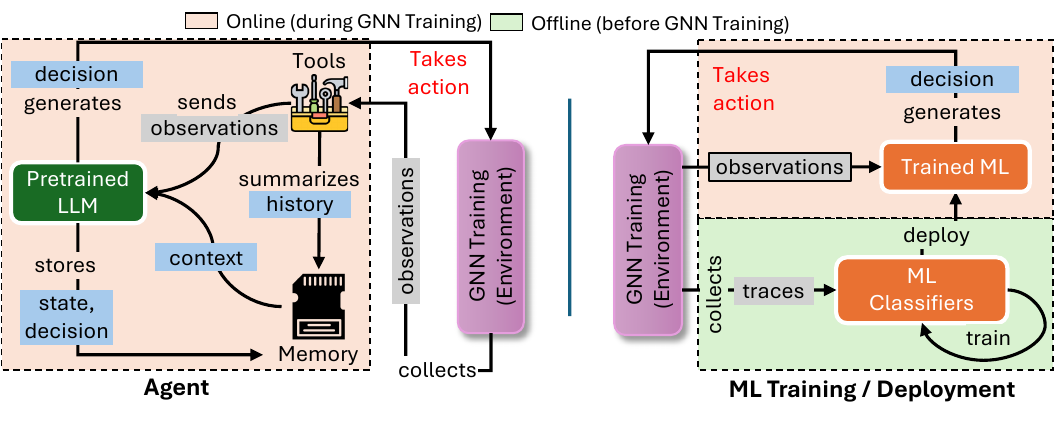}
  \caption{LLM agent learns by interacting with environment through auxiliary tools, whereas ML classifiers are trained offline.}
  \label{fig:agent}
\end{figure}

In Reinforcement Learning, a learning agent continuously observes an environment, learning from its own experiences, selecting actions that influence future observations~\cite{sutton1998reinforcement}. Modern LLMs extend this notion to \emph{language agents} that use an LLM as the core policy, but wrap it in an interaction loop with tools, memory, and external state to perform dynamic multi-step tasks autonomously~\cite{yao2022react, wang2024survey}. LLM agents are usually built from off-the-shelf instruction-tuned LLMs with no retraining (gradient updates), all adaptation happens via In-Context Learning (ICL). In contrast, the ML classifiers must be trained offline on collected traces before they can be deployed (as compared to LLM agents as shown in Fig.~\ref{fig:agent}).

\subsubsection{Tradeoffs}\label{ssec:qperf-tradeoff}
Key observations for distributed-memory GNN training are below. 
\begin{remark}[Strong scaling]\label{tradeoff-item:strong-scaling} 
     \#minibatches processed by a specific trainer reduces with increasing the resources (i.e., \#GPUs) across training epochs (assuming a fixed batch size).
\end{remark}
\begin{remark}[Diminishing overlap]\label{tradeoff-item:dim-overlap}
GNN model training on the current minibatch can be asynchronously overlapped with future minibatch preparation; additional tasks (i.e., prefetching\slash replacement) must also overlap with data-parallel training. However, with a reduction in \#minibatches (i.e., \emph{strong scaling}), training costs are no longer predominant (rather communication rises), diminishing any returns from overlap.
\end{remark}
\begin{remark}[Distribution shifts]\label{tradeoff-item:dist-shift} 
Variabilities are likely when observations rely on the input structure (highly irregular) and underlying execution environment (amenable to network contention). Since minibatches contribute to the observations processed by agents\slash classifiers, when they become sparse, it affects the decision accuracy. Additionally, hyperparameters such as batch size impact communication. Thus, \emph{distribution shifts} (deviation from training data) are probable. 
\end{remark}
We formalize these observations into approximate costs (in terms of the execution time, denoted as $T$) and decision quality, for ML classifiers (Supervised Learning (SL)) and LLM agents (In-Context Learning (ICL)).
\begin{corollary}[LLM agents require less resources for bootstrapping]
For nonzero observations set $\mathbb{S}_{\neq 0}$ and computational resources $\mathbb{R} \propto \{time\, , memory\}$, $\mathbb{R}^{\mathbb{S}}_{SL} > \mathbb{R}^{\mathbb{S}}_{ICL}$ considering comparable test times (i.e., $T_{test(\Theta=SL)} \simeq  T_{test(\Theta=ICL)}$).
\end{corollary}
We call pretraining costs to bootstrap LLM agent\slash ML classifier ($\Theta$) as \emph{offline} and costs during GNN training as \emph{online}. 
\begin{enumerate}[label=(\alph*),leftmargin=*]
\item \emph{Offline}: Collection of observations for training entails deploying the distributed GNN workload in \emph{trace-only} mode (training disabled, i.e., no backpropagation\slash optimization, weights frozen) to record per-minibatch sampling (takes $T_{Sampling}$ time) and buffer states across a variety of input\slash workload combinations, collecting $\mathbb{S}$ labeled samples, requiring $|\mathbb{S}| \times T_{Sampling}$. Training ($\Theta$) on these labeled datasets $\mathbb{S}$ (features from the sampler\slash buffer; labels are the replacement decisions) takes $T_{train(\Theta)}$, using supervised learning.
\item \emph{Online}: Cost of LLM agent\slash ML classifier decision is $T_{train(\Psi)} | T_{test(\Theta)}$, `$|$' indicates overlapping GNN ($\Psi$) training with agent\slash classifier's inference.

The overall execution time therefore is derived in terms of the respective offline and online components, considering training epochs ($e$) of minibatches ($\mathbb{M}$):
\begin{align}\label{eqn:cost_sl}
T_{SL} = \underbrace{\mathbb{S}\times(T_{Sampling}+T_{train(\Theta)})}_{\text{offline}}+\underbrace{\mathbb{M}\times e \times (T_{train(\Psi)}|T_{test(\Theta)})}_{\text{async online}}
\end{align}
Since LLMs exhibit ICL, there is no offline component (i.e., $\mathbb{S}_{= 0}$, no data preparation or training). 
\begin{align}\label{eqn:cost_icl}
T_{ICL}=\mathbb{M}\times e \times (T_{train(\Psi)}|T_{test(\Theta)})
\end{align}
\end{enumerate}
\begin{corollary}[LLM agents are resilient to distribution shifts]\label{lemma:decision-quality}
Contexts in training ML classifier (distribution $\mathbb{P}$) can be misaligned with test-time distribution ($\mathbb{Q}$), i.e., $\mathbb{P} \not\approx \mathbb{Q}$. 
\end{corollary}
    ML classifier learns a policy $\pi$ trained on $\mathbb{S}$ labeled samples from $\mathbb{P}$, with a loss of $\mathcal{L_\mathbb{P}}(\pi)$. LLM agent via ICL instead learns on $\mathbb{U}$ labeled samples from $\mathbb{Q}$ (where $|\mathbb{U}| < |\mathbb{S}|$) and learns policy $\pi'$ with loss $\mathcal{L_\mathbb{Q}}(\pi')$. Since GNN neighbor sampling is non-deterministic, $\mathbb{P} \not\approx \mathbb{Q}$. Under distribution shift, $\Delta=|\mathcal{L_\mathbb{Q}}(\pi)-\mathcal{L_\mathbb{P}}(\pi)| \gg 0$, and $\mathbb{U} \cap \mathbb{S} = \emptyset$. As long as the LLM agent's prediction error is $<\Delta$, then  $\mathcal{L_\mathbb{Q}}(\pi')<\mathcal{L_\mathbb{Q}}(\pi)$, i.e., it will be more accurate than ML classifier. 


\subsubsection{LLM characteristics}\label{sssec:llm_characteristics} Although LLMs with hundreds of billions (B) of parameters are known to be proficient in multifarious logical tasks and can activate fewer parameters during inference for optimizing the response times (e.g., Mixture-of-Experts routing, MoE~\cite{shazeer2017outrageously}), associated memory consumption is still proportional to the full parameters (roughly 2GB of memory is required every billion parameters, considering the half precision storage format). Remote self-hosting of the models is also prohibitive due to nontrivial network latencies, and despite declining costs, cloud-hosted models (i.e., Inference-As-A-Service) are often rate-limited~\cite{liagkou2024cost} (typically, thousands of tokens\slash minute, too restrictive for our usage). We prefer openly available LLMs with \emph{small-medium parameters} (ideally fewer than 5B and quantized to further reduce memory) and medium context lengths, for computational efficiency reasons (minimum end-to-end response time and GPU memory residency). Particularly, we narrow down to consider small\slash medium-sized quantized LLMs that exhibit high scores on popular evaluation benchmarks for \emph{mathematical reasoning} (as measured by the \textsc{MATH-500}~\cite{lightman2023let} benchmark) and \emph{instruction compliance} (for adhering to predefined question-answer formats, covered by \textsc{IFEval}~\cite{zhou2023instruction}). 
\begin{figure}[t]
  \centering
  \includegraphics[width=\linewidth]{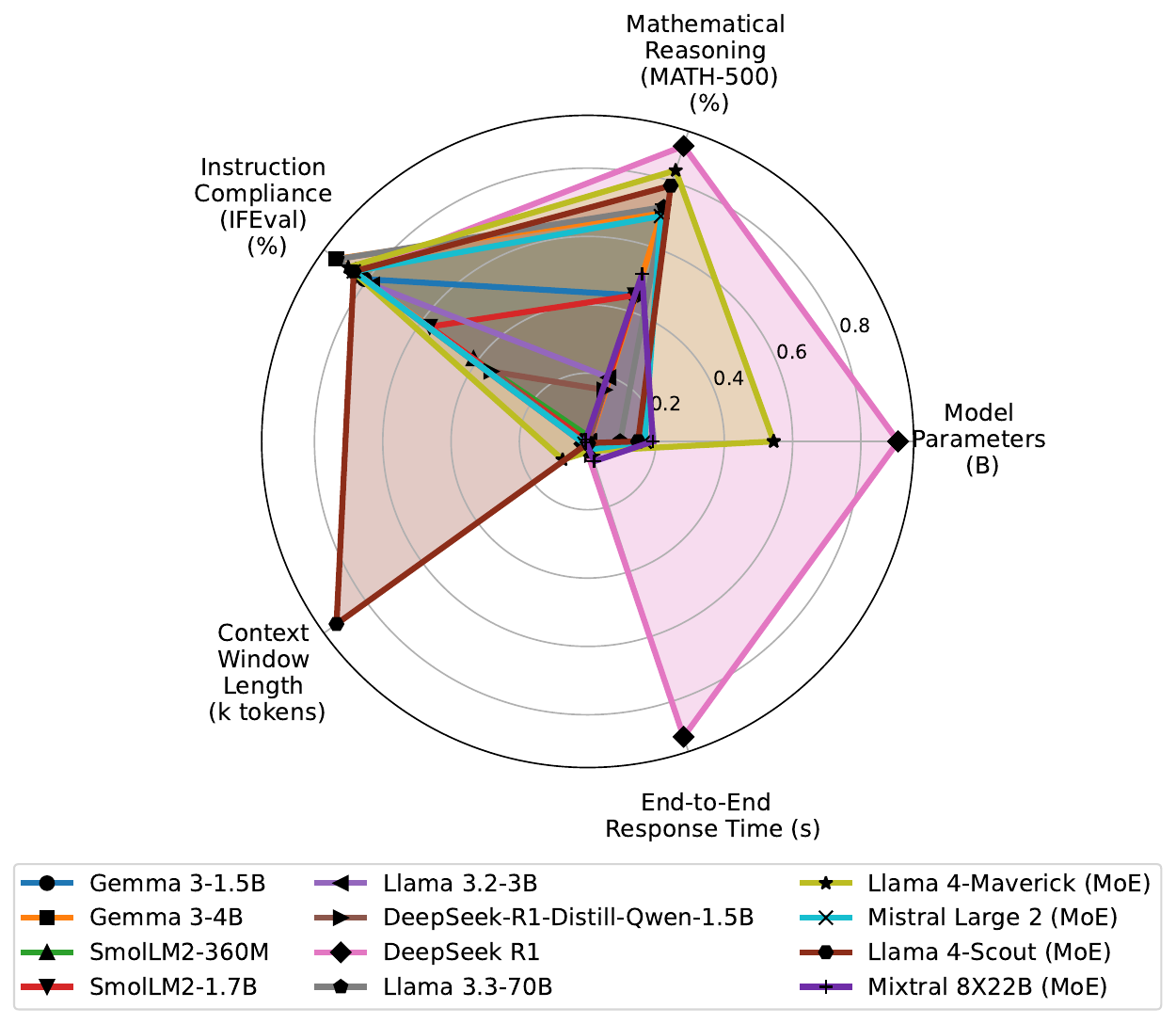}
  \caption{Quantifying LLMs on key characteristics.}
   \label{fig:spider-llms}
\end{figure}
Our choices are quantified in the spider chart on Fig.~\ref{fig:spider-llms} in terms of high-performance\slash memory-usage and problem-solving\slash instruction-following capabilities.

\section{Related Work}\label{sec:related}

Supervised learning (SL) is widely applied for system performance optimization~\cite{wu2022survey}, including performance prediction, memory and cache management~\cite{10.1145/3611018, shi2019applying, 8668490}. Models are typically trained offline on labeled traces or simulation data and then deployed as fixed policies at test time. Recently, LLMs have been explored for system-level tasks: anomaly detection \cite{jin2024large}, compiler autotuning~\cite{cummins2023large}, and task mapping \cite{niu2024fair}. 
In-context learning (ICL)~\cite{brown2020language} and \textit{chain-of-thought prompting}~\cite{wei2022chain} allow LLMs to adapt to new tasks without gradient updates by reasoning over a few examples directly in prompt, unlike traditional ML models that must be retrained. This has motivated a growing body of work that uses LLMs as optimizers ~\cite{li2022optformer,yang2023large,chen2023evoprompting,lin2024llm}. Recent works have gone beyond passive prediction by utilizing LLMs as active agents that search extremely large design spaces with only bandit-style feedback \cite{sutton1998reinforcement, bubeck2012regret}. For e.g., AgentHPO~\cite{liu2025agenthpo} in hyperparameter optimization, automatic routing solver~\cite{li2025ars} in vehicle routing problem, chemical process optimization~\cite{zeng2025llm}, cluster management~\cite{vadisetty2025ai}, operations research optimization~\cite{zhang2025or}, wireless resource and power allocation~\cite{tong2025wirelessagent}, multi-robot task planning~\cite{kannan2024smart}, GPU kernel optimization~\cite{dong2025stark}, and SIMD vectorization of HPC kernels~\cite{taneja2025llm}. Beyond optimization, LLM agents can also plan and adapt over time in unfamiliar environments~\cite{shen2023hugginggpt, schick2023toolformer, park2023generative, lin2023swiftsage, shinn2023reflexion, zhao2024expel, ahn2022can}, demonstrating efficient decision-making. While early work focused on large models, recent evidence suggests small language models (SLMs, <10B parameters) can match or beat larger LLMs on many benchmarks at much lower cost~\cite{belcak2025small}. This is particularly relevant in systems and HPC, where agents must coexist with primary workloads on shared accelerators.

Unlike the adaptive replacement strategy proposed by \emph{Rudder}, existing static\slash heuristics-driven prefetching strategies to reduce data movement in shared\slash distributed memory GNN training~\cite{liu2023bgl, yang2019aligraph, lin2020pagraph, kaler2023communication} exhibits parameter-tuning and preprocessing overheads.

\section{Enhancing GNN Training using \emph{Rudder}}\label{sec:method-design}
\begin{figure}[h]
  \centering
  \includegraphics[width=\linewidth]{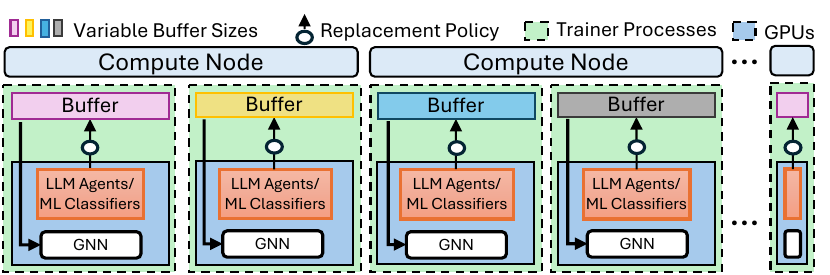}
  \caption{Agents\slash classifiers deployed alongside GNN training tasks, trainer processes access persistent buffers.}
 \label{fig:llm-deploy}
\end{figure}
In this section, we detail the LLM/ML execution environment (\S\ref{ssec:method-llm-exec}), agentic prefetching workflow (\S\ref{ssec:method-workflow}), agent and classifier decision making (\S\ref{ssec:method-llm-decision}, \S\ref{ssec:method-non-llm}), synchronous\slash asynchronous variants (\S\ref{ssec:method-dist-approach}), and evaluation criteria (\S\ref{ssec:method-eval}).
\subsection{Tasks Creation and Deployment}\label{ssec:method-llm-exec}

\begin{figure}[h]
  \centering
  \includegraphics[width=\linewidth]{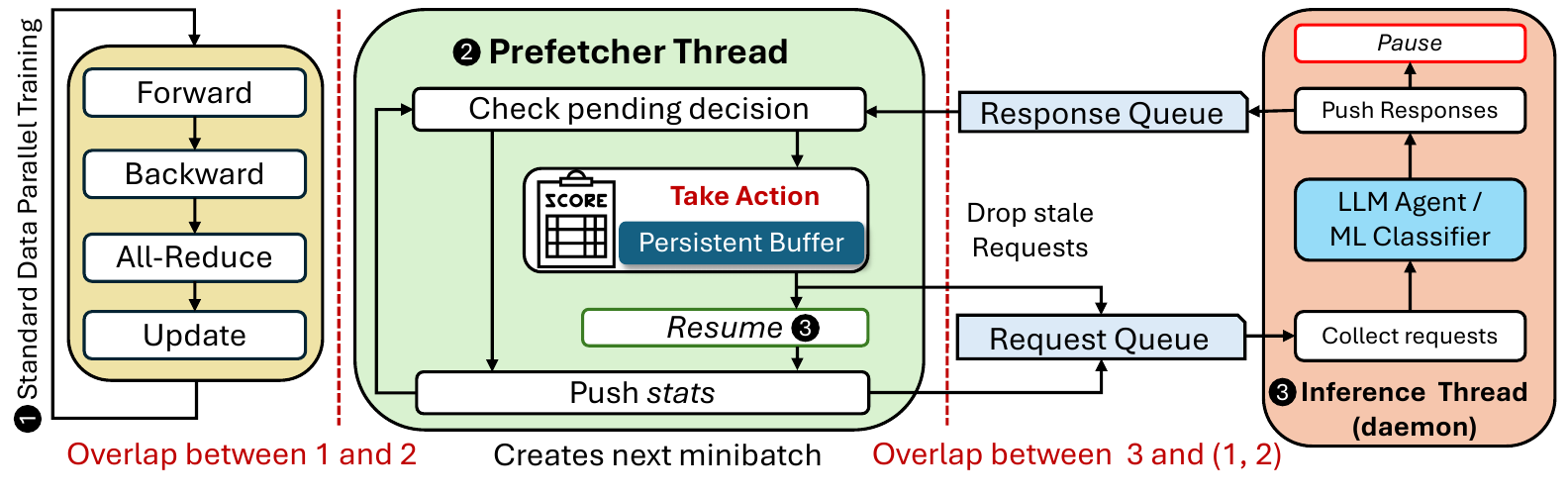}
  \caption{Tasks in the trainer processes of \emph{Rudder}.}
 \label{fig:daemon}
\end{figure}

Adaptive replacement includes two concurrent tasks: \emph{prefetching}, which owns and enforces the scoring policy (Fig.~\ref{fig:scoring}), and persistent buffer replacement which decides \emph{when} to replace via LLM agents/ML Classifiers (Fig.~\ref{fig:agent}). After loading the graph partitions into memory, each trainer process asynchronously offloads the prefetching task to CPU threads and runs the inference task in a background \emph{daemon} thread (Fig.~\ref{fig:daemon}). Both inference and primary GNN training tasks share the same GPU (as shown in Fig.~\ref{fig:llm-deploy}), relying on a lightweight inference engine for processing requests. We use a combination of Python \textsc{ThreadPoolExecutor}~\cite{threadpool} and NUMBA~\cite{lam2015numba} to perform prefetching tasks, utilizing asynchronous and parallel execution with CPU threads. Inference tasks are spawned using Python’s \texttt{threading} module~\cite{threading}. We deploy LLM inference via \textsc{ollama}~\cite{ollama} (front-end to popular \texttt{llama.cpp} project~\cite{llamacpp}, a C++-based implementation with minimum external dependencies using vendor-optimized GPU kernel implementations) inside the daemon thread. The daemon shares request\slash response queues with the prefetcher to exchange runtime metrics and suggested actions.

\subsection{Components of the Agentic Workflow}\label{ssec:method-workflow}
\begin{figure}[h]
  \centering
  \includegraphics[width=\linewidth]{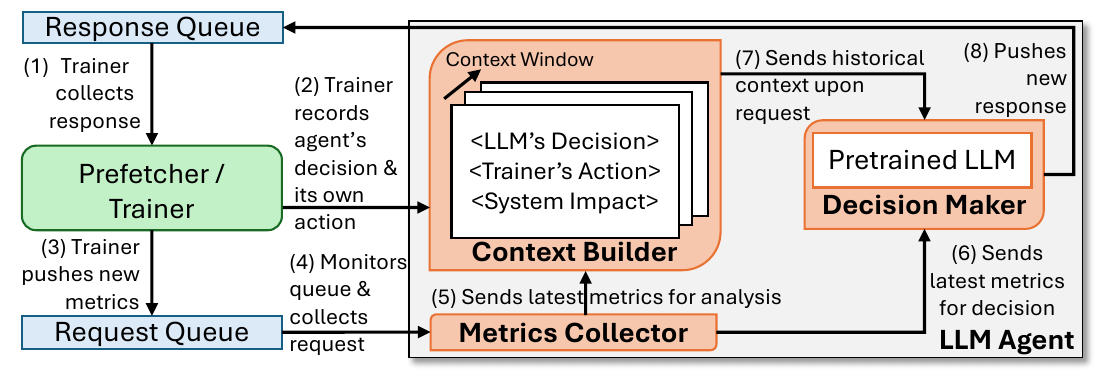}
  \caption{LLM agent request\slash response flow and context updates.}
 \label{fig:workflow}
\end{figure}
We now explain Rudder's LLM agent workflow. As shown in Fig.~\ref{fig:workflow}, the trainer and LLM agent coordinate via shared request/response queues: the trainer enqueues runtime observations and consumes the returned action. The LLM agent comprises three components that orchestrate decision-making throughout the training.
The associated events are listed numerically in Fig.~\ref{fig:workflow} to indicate the sequence of actions taken by each component and prefetcher tasks at a given time.
\begin{itemize}[leftmargin=*,nosep]
\item \textsc{Metrics Collector}: This component continuously streams key execution metrics (e.g., \%-Hits, remote communication volume, and minibatch progress details) to the LLM, providing \emph{temporal context} to reason about the potential replacement benefit in the persistent buffer versus the associated communication costs (steps 5 and 6 in Fig.~\ref{fig:workflow}), while accounting for remaining training epochs. 
\item \textsc{Context Builder}: Operating alongside the \textsc{Metrics Collector}, this component tracks past replacement-related events and their outcome. When a replacement decision is executed or skipped, the \textsc{Context Builder} captures the pre-decision metrics and, upon availability of the next set of metrics, evaluates the previous decision's \emph{effectiveness}, maintaining a history of LLM's decisions and their impact, providing \textit{context} for the future (step 7 in Fig.~\ref{fig:workflow}).
\item \textsc{Decision Maker}: Using the insights from the \textsc{Metrics Collector} and \textsc{Context Builder}, the local LLM decides whether an \emph{action} (replacement of nodes) must be taken. The \textsc{Decision Maker} constructs a comprehensive context, which combines the static (graph and training metadata) and dynamic information (recent metrics and the replacement history) to formulate the task and stores responses back in the queue for the trainer (step 8 in Fig.~\ref{fig:workflow}).

\end{itemize}

\subsection{LLM Agent Decision Making}\label{ssec:method-llm-decision}
Rudder continuously shares local execution\slash configuration metrics with the LLM, classified herein.
\begin{itemize}[leftmargin=*,nosep]
    \item \textit{Persistent buffer}: \%-Hits (i.e., percent of sampled remote nodes present in the local persistent buffer), and actual number of nodes replaced (as \% of the buffer size).
    \item \textit{Training}: Communication volume (\#remote nodes fetched), and current/pending \#minibatches for \emph{progress awareness} (avoid replacements near completion).
    \item \textit{Replacement history}: Impact of past replacement decisions provided by \textsc{Context Builder} (changes in \%-Hits and communication volumes).
    \item \textit{Graph structural information (static)}: The number of vertices\slash edges in the graph and in the local partition.
\end{itemize}
In \S\ref{sssec:method-llm-trajectory}, we discuss LLM decision-making process, followed by contextualizing the input tasks into ``prompts'' in \S\ref{sssec:method-llm-prompt}.
\begin{figure}[h]
  \centering
  \includegraphics[width=0.9\linewidth]{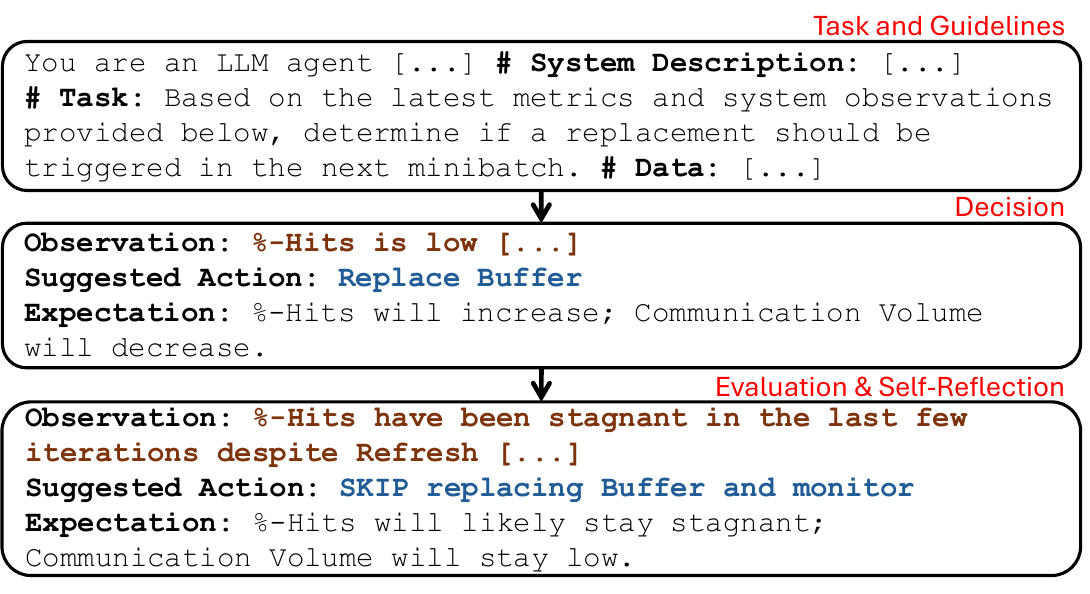}
  \caption{LLM agents making decisions.}
 \label{fig:prompt}
\end{figure}
\subsubsection{\textbf{Decision Trajectory}}\label{sssec:method-llm-trajectory}
The LLM is prompted with a structured task definition including the system's description, task objective, and relevant context. The task requests the LLM to determine whether to trigger a replacement for the next minibatch based on the latest observations (i.e., execution metrics). Based on the observations provided (e.g., low \%-Hits), the LLM suggests an action (e.g., refreshing the persistent buffer to increase \%-Hits, reducing the communication). The LLM also formulates an expectation for the outcome, to evaluate the effectiveness of the decision in subsequent iterations, and to adjust its strategy. For instance, if \%-Hits remains unchanged, the LLM might suggest skipping further replacements (e.g., in Fig.~\ref{fig:prompt}: decision $\rightarrow$ evaluation).
\subsubsection{\textbf{Prompt Engineering}}\label{sssec:method-llm-prompt}
In \textit{zero-shot} ICL prompting, task-specific instructions are provided without any examples and the model relies entirely on its pre-existing knowledge and reasoning capabilities. Our prompt (partially illustrated in Fig.~\ref{fig:prompt}) for the LLM's decisions explicitly provides structured context (explaining the \emph{meaning} and \emph{importance} of metrics) designed to clearly communicate the current state of the buffer, historical replacement effectiveness, and relevant graph metadata. 
Chain-of-Thought (CoT)~\cite{wei2022chain} improves quality (e.g., higher \%-Hits) at 4–5$\times$ response latency.
\subsection{ML Classifier Decision Making}\label{ssec:method-non-llm}
Aside from LLMs, Rudder can optionally deploy supervised ML classifiers that consumes the same execution metrics and returns a binary decision. We train the models offline on execution traces (e.g., \%-Hits, communication latency, buffer occupancy) collected across several datasets, partition configurations, and buffer sizes to expose them to diverse setups, using hundreds to thousands of node-hours additionally.
In pretraining, since execution traces are unlabeled, we assign labels by comparing the key metrics before and after replacement events. For successive minibatches, we capture the relative changes in \%-Hits ($\Delta\%Hits$) and communication costs associated with retrieving remote node features ($\Delta T_{COMM}$). A replacement instance is labeled ``good'' if the improvement in \%-Hits outweighs increase in communication latency: $S'=\Delta\%Hits -\Delta T_{COMM}>0$, otherwise ``bad'' ($S'=0$). Scenarios exist that compromise the integrity of the labels: (i) sampling\slash communication are prone to variations and delayed effects (ii) stateless inference inhibits reasoning over context histories or causal chains (e.g., previous eviction reduced communication but did not raise \%-Hits) (iii) insufficient pretraining due to the massive search space of the execution configurations (\S\ref{ssec:results-finetune}). Hence, Rudder optionally uses \textit{online fine-tuning} mechanism to periodically update the model's decision head while keeping the weights frozen. 
\subsection{Distributed-Memory Implementation}\label{ssec:method-dist-approach}
Rudder asynchronously offloads prefetching and inference tasks (Fig.~\ref{fig:threads} and \S\ref{ssec:method-workflow}) with protected shared queues to efficiently process the next minibatch, overlapping with data-parallel training loop on the current minibatch.
Concurrently, the prefetcher thread pushes the latest minibatch metrics onto the request queue and regularly checks for incoming decisions on the response queue (Fig.~\ref{fig:workflow}). Since, Rudder does not alter the underlying sampling algorithm or the data-parallel training, training accuracy remains intact.

\subsubsection{\textbf{Asynchronous vs. Synchronous}}\label{ssec:method-syncasync} The time taken for an inference model to process input metrics and return a decision is the \emph{response time}. For LLMs, this includes the latency of the \textsc{Ollama} server instance (see \S\ref{ssec:method-llm-exec}) to process a ``prompt'' and return a response, whereas for ML classifiers it corresponds to their inference time.
In practice, the inference model's decision time can exceed that of the trainer processing the subsequent minibatch, causing ``stale'' requests in the shared request queue (returned decision is invalid due to obsolete metrics), disrupting the decision-making pipeline. To avoid this particular issue (i.e., a variant of classic \emph{producer-consumer} problem), after a decision is placed in the response queue, the inference thread \emph{pauses} itself and is only \emph{resumed} by the prefetcher thread once the decision is processed and the queues are cleared of backlog. 
\begin{figure}[t]
  \centering
  \includegraphics[width=\linewidth]{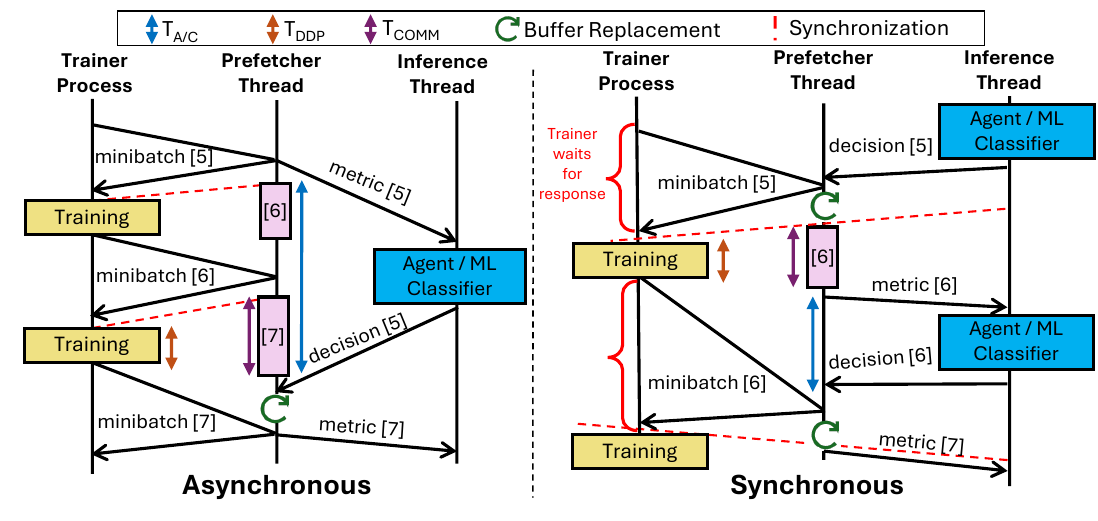}
  \caption{Asynchronous (default) and Synchronous (trainer waits for agent’s response) modes of Rudder.}
 \label{fig:threads}
\end{figure}
A side-effect of this design, mainly in LLM agents, is that in some minibatches, there may not be any current context for the LLM after clearing the shared queues (past contexts still exist). This is the default execution scenario of Rudder, referred to as the \emph{asynchronous} version. To prevent the loss of intermediate context due to ``stale'' requests from past minibatches, prefetcher and inference tasks can be synchronized; this \emph{synchronous} version does not suffer from ``stale'' requests and may exhibit better accuracy, but is impractical due to coarse-grain synchronizations. Fig.~\ref{fig:threads} demonstrates the variants---the prefetcher wakes up the inference thread before sending the metrics associated with minibatches 5 and 7 for the asynchronous version (skipping minibatch \#6), whereas for the synchronous version, every minibatch is present in the request-response queue. This ``gap'' between consecutive replacement events is referred to as the \emph{replacement interval} ($r$); $r\geq$1 for the asynchronous (higher numbers imply agent overheads), and $r=1$ for the synchronous variant.

\subsubsection{\textbf{Distributed-Memory Algorithm}}\label{ssec:method-dist-alg}
The steps in the default asynchronous variant of Rudder are explained in Algorithm~\ref{alg:producer_prefetch}. After initializing the inference thread to run in the background, the standard training loop is invoked between Lines~\ref{alg:begin_training}--\ref{alg:end_training}. After obtaining the latest minibatch from queue (in Line~\ref{alg:get_minibatch} of Algorithm~\ref{alg:producer_prefetch}), the data-parallel training can begin (Line~\ref{alg:ddp_training}), while the prefetcher thread asynchronously samples the next minibatch (Line~\ref{alg:nbr_sampling}), checking the remote nodes in the persistent buffer (Line~\ref{alg:next_minibatch}), and retrieving the replacement decision (from the shared response queue, Line~\ref{alg:check_decision}) if available. If the decision is to replace the nodes, then nodes in the persistent buffer whose scores are below a threshold (i.e., $0.95$, see \S\ref{ssec:motiv-replacement}) are queued for communication (utilizing DistDGL's Remote Procedure Call or RPC framework~\cite{zheng2020distdgl} which performs sender-side aggregation and multithreaded point-to-point communication over TCP/IP Sockets API), as depicted in Line~\ref{alg:replace_fetch}. Then, the shared request queue is cleared in Line~\ref{alg:clear_requests} before notifying the inference thread (to ensure prefetcher shares the latest runtime metrics, Line~\ref{alg:resume_thd}). As shown in Line~\ref{alg:queue_minibatch}, before exiting, the prefetcher thread queues the next minibatch for data-parallel training, initiating communication for sampled nodes (missing from the persistent buffer). 
\begin{algorithm}
{\footnotesize
\caption{\footnotesize \textsc{\textbf{Distributed GNN Training}} \\
\textbf{Inputs:} $S$: Sampled Nodes, $BUF$: Persistent Buffer, $\Theta$: Pretrained LLM / ML Classifier, $G_P$: Graph Partition, $Q_M$: Minibatch Queue, $Q_R, Q_{\tilde{R}}$: Agent's Request\slash Response Queue, $\Pi$: Scoring Policy  \\
\textbf{Output:} $\Psi$: Trained GNN Model
}
\label{alg:producer_prefetch}
\begin{algorithmic}[1]
    \State Initialize GNN model $\Psi$, $BUF$
    \State Spawn \Call{InferenceThread}{} as \textit{daemon} \label{alg:init_agent}
    \For{$e$ in \textit{range(\texttt{\#epochs})}} \label{alg:begin_training}
        \For{$step$ in \textit{range(\texttt{\#minibatches})}}
        \State \(\textit{minibatch} \gets {Q_M}.get()\) \Comment{Extract minibatch from queue} \label{alg:get_minibatch}
        \State \textbf{\texttt{async}} \Call{PrefetcherThread}{$Q_M$} 
        \State \Call{TorchDDP}{$\psi$, \textit{minibatch}} \Comment{Data-Parallel Training} \label{alg:ddp_training}
        \State \textbf{\textsc{Synchronize}} \label{alg:end_training} \Comment{Gradient sync. across trainers}
        \EndFor
    \EndFor
    \Procedure{PrefetcherThread}{$Q_M$}
    \State \(S \gets \Call{NeighborSampler}{G_P}\) \Comment{Sample remote nodes} \label{alg:nbr_sampling}
    \State \(\textit{minibatch} \gets (BUF \cap S)\) \Comment{Copy nodes from buffer} \label{alg:next_minibatch} 
    \State \(\textit{decision} \gets {Q_{\tilde{R}}.get()}\) \Comment{Check for response (Non-Blocking)} \label{alg:check_decision}
    \If{\(\textit{decision}\) is found and \(\textit{decision} = Replace\)} \label{alg:yes_replace}
        \State \(\Call{ReplaceandFetch}{BUF, \Pi}\) \Comment{Replace nodes, fetch remote} \label{alg:replace_fetch}
        \State \({Q_R}.clear()\) \Comment{Clear stale requests} \label{alg:clear_requests}
        \State \textsc{\textbf{Notify}} \textsc{InferenceThread}$(Q_R,Q_{\tilde{R}})$ \Comment{Release lock} \label{alg:resume_thd}
    \EndIf 
    \State \(missed\_nodes \gets (S\setminus BUF)\) \Comment{Missing in buffer, fetch remote} \label{alg:missed_nodes}
    \State \(Mr \gets\) runtime metrics \Comment{Generate runtime metric} \label{alg:gen_metrics}
    \State \({Q_R}.put(M_r)\) \Comment{Put metrics in the queue for LLM Agent/ML Classifier} \label{alg:insert_metrics}
    \LComment{Put minibatch in the queue for next DDP} 
    \State \({Q_M}.put(minibatch+\Call{Fetch}{missed\_nodes})\) \label{alg:queue_minibatch}
    \EndProcedure
    \Procedure{InferenceThread}{$Q_R,Q_{\tilde{R}}$} \label{alg:agent_thd_begin}
        \While{\textbf{true}} \Comment{Wait for current metric}
            \If{$\Theta=LLM$}
                \State \(M_r \gets \Call{MetricsCollector}{Q_R}\) \label{alg:call_metricagent}
                \State \(C_r \gets \Call{ContextBuilder}{M_r}\) \Comment{Update\slash get current Context} \label{alg:call_contextagent}
                \State \(decision \gets \Call{DecisionMaker}{M_r,C_r,\Theta}\) \Comment{Inference} \label{alg:call_decisionagent}
            \Else
                \State \(decision \gets \Call{DecisionMaker}{Q_R,\Theta}\) \Comment{Inference} \label{alg:call_disc_decision}
                \If{finetune} \Call{CollectAndFinetune}{$Q_R, \Theta$} \label{alg:finetune} \Comment{Optional} \EndIf
            \EndIf
            \State \(Q_{\tilde{R}}.push(decision)\) \label{alg:push_decision} \Comment{Put decision in queue}
            \State \textbf{\textsc{WaitUntilNotified}} \Comment{Wait for prefetcher to take action} \label{alg:pause_done}
        \EndWhile
    \EndProcedure
\end{algorithmic}
}
\end{algorithm}
The inference thread invokes the inference model to coordinate the request\slash response data streams as discussed in \S\ref{ssec:method-workflow}. The tasks performed within the inference workflow are shown between Lines~\ref{alg:agent_thd_begin}--\ref{alg:pause_done} in Algorithm~\ref{alg:producer_prefetch} (corresponding to Fig.~\ref{fig:workflow}). For LLMs (Lines~\ref{alg:call_metricagent}--\ref{alg:call_decisionagent}), the workflow expands into a multi-stage pipeline: the \textsc{MetricsCollector} first aggregates raw metrics (\%-Hits, communication volume, etc.), which are passed to the \textsc{ContextBuilder} to maintain history before being consumed by the \textsc{DecisionMaker} for inference. The classifier executes a direct stateless inference step on the current metrics (Line~\ref{alg:call_disc_decision}), bypassing context management, with optional periodic finetuning on buffered minibatches (Line~\ref{alg:finetune}). In both cases, the resulting decision is pushed into the shared queue (Line~\ref{alg:push_decision}), before receiving a notification (i.e., releasing a lock for the prefetcher thread, and waiting to re-acquire it~\cite{threadinglock}). 

\subsubsection{\textbf{Performance Modeling}}\label{ssec:method-perfmodel} 
In general, the overall performance can be determined by the data-parallel training, agent/classifier inference, and associated communication\slash synchronization overheads, denoted as $T_{DDP}$, $T_{A/C}$, and $T_{COMM}$, respectively. Other essential overheads pertaining to replacement activities, such as persistent buffer processing, tallying nodes for replacement, searching for remote nodes in buffer, and other similar operations, are relatively mild in comparison. For simplicity, we exclude periodic fine-tuning overheads for classifiers.
For the asynchronous variant of Rudder (default), the trainer is overlapped with the prefetcher, and the inference task is overlapped with both of these activities---execution time is proportional to $\mathit{max}(T_{DDP}, T_{COMM}) / T_{A/C}$.
Conversely, the execution time performance of the synchronous variant can be expressed as: $T_{DDP} / (T_{A/C} + T_{COMM})$. Although there is some overlap between the trainer and prefetcher tasks working on disparate minibatches, the trainer must wait until the prefetcher and inference have processed the current minibatch. Unless data-parallel training overheads are significant (i.e., $T_{DDP} >> (T_{COMM} + T_{A/C})$, Remark~\ref{tradeoff-item:dim-overlap}), no performance can be derived. This is also unlikely as the \#minibatches is inversely proportional to the \#trainers, assuming constant batch sizes (Remark~\ref{tradeoff-item:strong-scaling}). 

\subsection{LLM Evaluation for Prefetching}\label{ssec:method-eval}
Establishing ground-truth or reference solution to assess the \emph{functional correctness} of an action (i.e., labeling a decision as \textit{pass} or \textit{fail}) is challenging. Therefore, we develop an unconventional \emph{reference-free functional correctness check} to evaluate agentic prefetching based on empirical observations. We consider LLM's predicted impact on the system as a self-consistency check. Once the agent takes an action \(a_{t}\) at a given step $t$, the environment transitions to the observed state \(s_{t+1}\), and we compare it against the LLM's predicted state \(\hat{s}_{t+1}\). Alignment between \(s_{t+1}\) and \(\hat{s}_{t+1}\) reaffirms that the agent made a sound decision and accurately anticipated its effect (i.e., a ``pass''), whereas substantial deviations imply suboptimal or misguided reasoning (i.e., a ``fail''). For instance, if the LLM predicts that replacing nodes at a given step will increase the \%-Hits, but in reality no improvement is observed, we deem that the decision is a \emph{fail}. We formalize this reflection-based approach by adopting a popular metric for evaluating functional correctness; we use the \textit{Pass@K} ($K=1$)~\cite{chen2021evaluating} metric on \%-Hits, measuring how often the real-world changes match with the agent's predictions.


\section{Evaluation and Analysis}\label{sec:results}
\begin{figure*}[!ht]
  \centering
  \includegraphics[width=\linewidth]{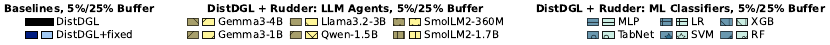} 
  \begin{subfigure}[b]{\linewidth}
      \centering
      \includegraphics[width=0.9\linewidth]{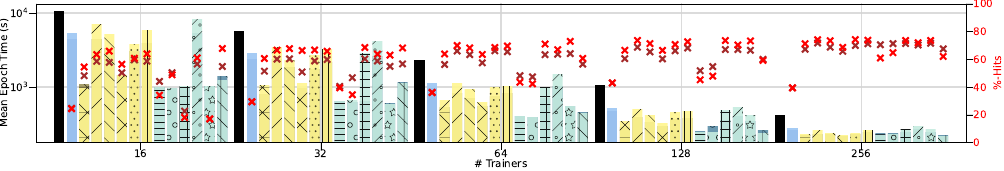}
      \vspace{-0.2cm}
      \caption{papers100M}
      \label{fig:papers-scale}
  \end{subfigure}
  \begin{subfigure}[b]{0.45\linewidth}
      \centering
      \includegraphics[width=\linewidth]{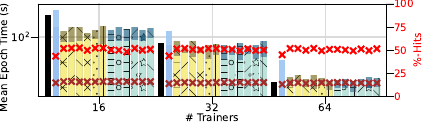}
      \vspace{-0.6cm}
      \caption{reddit}
      \label{fig:reddit-scale}
  \end{subfigure}
  \begin{subfigure}[b]{0.45\linewidth}
      \centering
      \includegraphics[width=\linewidth]{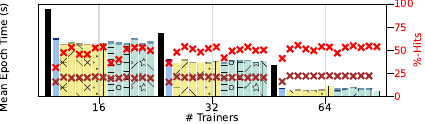}
      \vspace{-0.6cm}
      \caption{orkut}
      \label{fig:orkut-scale}
  \end{subfigure}
  \begin{subfigure}[b]{0.45\linewidth}
      \centering
      \includegraphics[width=\linewidth]{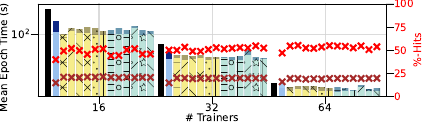}
      \vspace{-0.6cm}
      \caption{products}
      \label{fig:products-scale}
  \end{subfigure}  
  \begin{subfigure}[b]{0.45\linewidth}
      \centering
      \includegraphics[width=\linewidth]{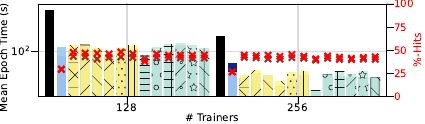}
      \vspace{-0.6cm}
      \caption{friendster}
      \label{fig:friendster-scale}
  \end{subfigure}
  \caption{DistDGL+Rudder's performance with agents and classifiers across datasets, \#trainers, and persistent buffer sizes (5\%/25\%). The bars show mean epoch time (left Y; \emph{lower} is better; 5\%/25\% bars are overlapped), with \%-Hits (\textcolor[HTML]{8B4513}{$\bm{\times}$}=5\%\slash \textcolor[HTML]{FF0000}{$\bm{\times}$}=25\%) (right Y; \emph{higher} is better).}
  \label{fig:combined_scale}
\end{figure*}
\begin{table}[t]
\centering
\scriptsize
\begin{threeparttable}
\caption{(a) Datasets and (b) Models used.} \label{tab:datasets-models}
\begin{tabular}{lccc}
\hline
\multicolumn{4}{c}{\textbf{(a) Datasets}} \\
\hline
\textbf{Dataset} & \textbf{Nodes $|V|$} & \textbf{Edges $|E|$} & \textbf{Feat. Dim.} \\
\hline
products~\cite{ogb2020}  & 2.4M   & 61.85M   & 100  \\
reddit~\cite{ogb2020}    & 0.23M  & 114.61M  & 602  \\
papers100M~\cite{ogb2020}    & 111M   & 1.6B     & 128  \\
orkut~\cite{snapnets}     & 3.07M  & 117.18M  & 8    \\
friendster~\cite{snapnets}& 65.6M  & 1.8B     & 128  \\
yelp~\cite{yelp}& 716K & 13.9M & 300 \\
ogbn-arxiv~\cite{wang2020microsoft}& 169K & 1.1M & 128 \\
\hline
\hline
\multicolumn{4}{c}{\textbf{(b) Characteristics of LLM Agents}} \\
\hline
\textbf{Model Name} & \textbf{Size (Model / KV Cache) (GB)} & \textbf{Quantization} & \textbf{Type} \\
\hline
Gemma3-4B~\cite{gemma3mc} & 3.3 / 0.27 & Q4\_K\_M & Base \\
Gemma3-1B~\cite{gemma31bmc} & 0.8 / 0.05 & Q4\_K\_M & Base\\
Llama3.2-3B~\cite{llama3mc} & 2 / 0.22 & Q4\_K\_M & Base\\
SmolLM2-360M~\cite{smollm2mc} & 0.38 / 0.08 & Q4\_K\_M & SLM\\
SmolLM2-1.7B~\cite{smollm21bmc} & 1.06 / 0.38 &  Q4\_K\_M & SLM\\
Qwen-1.5B\tnote{$\star$}\hspace{0.2cm}\cite{qwenmc}  & 10 / 0.05 &  F16 & Distill\\
\hline
Mixtral-8x7B~\cite{mixtral7bmc}  & 24 / 0.26 &  Q3\_K\_L & MoE\\
Mixtral-8x22B~\cite{mixtral22bmc} & 52 / 0.45 &  Q2\_K & MoE\\
Granite-3.1-3B~\cite{granite3mc} & 6.6 / 0.13 & F16 & MoE \\ 
\hline 
\end{tabular}
\begin{tablenotes}
\footnotesize
\item[$\star$]{DeepSeek-R1-Distill-Qwen-1.5B.}
\end{tablenotes}
\end{threeparttable}
\end{table}   

We introduce the experimental variants and platforms\slash datasets\slash models, then examine baseline performance\slash scalability (\S\ref{ssec:results-training-performance}), performance\slash persistence tradeoffs (\S\ref{ssec:results-buffer-sizes}), synchronous vs. asynchronous variants (\S\ref{ssec:results-latency}), performance on unseen data (\S\ref{ssec:results-finetune}), replacement trajectory (\S\ref{ssec:results-llm-reasoning}), and MoE agents (\S\ref{ssec:results-moe}). We evaluate the following \textbf{variants}:
\begin{enumerate}[leftmargin=*,nosep]
    \item \textbf{DistDGL} (\emph{no prefetch\slash overlap}): Baseline DistDGL, for every sampled minibatch, trainer PEs fetch remote nodes.
    \item \textbf{DistDGL+fixed} (\emph{static prefetch w overlap}): Concurrent minibatch processing and replacement decisions at \textit{every} minibatch (reasonable middle-ground, see \S\ref{ssec:motiv-replacement}).
    \item \textbf{DistDGL+Rudder} (\emph{dynamic prefetch w overlap}):  LLM agent\slash ML Classifier enabled dynamic replacement decisions (see \S\ref{ssec:motiv-suitability}) in addition to concurrent minibatch processing (i.e., \emph{Rudder}).
Both DistDGL+fixed and DistDGL+Rudder use the same scoring policy for determining replacement candidates (\S\ref{ssec:motiv-replacement}).
\end{enumerate}

\noindent {\textbf{Datasets\slash Models\slash Platform}}: 
    We use four OGB~\cite{ogb2020} datasets (papers, reddit, products, and arxiv), yelp~\cite{yelp}, and two SNAP~\cite{snapnets} social networks (orkut and friendster), performing node classification with a 2-layer GraphSAGE model (fanout $\{10, 25\}$, batch size $2000$). For SNAP datasets lacking node labels, we use \texttt{node2vec}~\cite{grover2016node2vecscalablefeaturelearning} and assign pseudo labels based on the top-5000 communities. Experiments use DistDGL~\cite{dgl} v2.5 (server\slash node: \#trainers = \#GPUs, graphs partitioned using METIS~\cite{karypis1998fast}), PyTorch v2.4.0 with CUDA 12.1, and NCCL v2.10.5. Generative inference runs on \textsc{Ollama}~\cite{ollama2025} v0.4.7. LLMs are from HuggingFace~\cite{huggingface}, quantized on group-wise granularity~\cite{gong2024survey}. Datasets and LLMs are summarized in Table~\ref{tab:datasets-models}. We fix the LLM context window (<2048 tokens), bounding KV-cache growth keeping the per-agent KV allocations modest across all LLMs (Table~\ref{tab:datasets-models}(b)). Aside from LLM agents, we also consider a mix of traditional and modern ML classifiers
    such as Multi-Layer Perceptron (MLP), Logistic Regression (LR), Decision Trees (XGBoost~\cite{chen2016xgboost} (XGB) and Random Forests (RF)), Support Vector Machines (SVM) and TabNet~\cite{arik2021tabnet} with sequential attention for feature selection.

Experiments are performed on NERSC Perlmutter supercomputer: 1,792 GPU nodes (64-core 2.4GHz AMD EPYC 7763 CPUs, 256GB DDR4 RAM, 256MB L3 cache, 4$\times$ NVIDIA A100 GPUs with 40GB HBM2; 256 nodes are equipped with 80GB HBM2e, exhibiting about 20\% higher bandwidth) and HPE Slingshot 12 interconnect~\cite{yang2020accelerate}.

\subsection{Baseline Performance}\label{ssec:results-training-performance}
\begin{figure}[!ht]
  \centering
  \includegraphics[width=\linewidth]{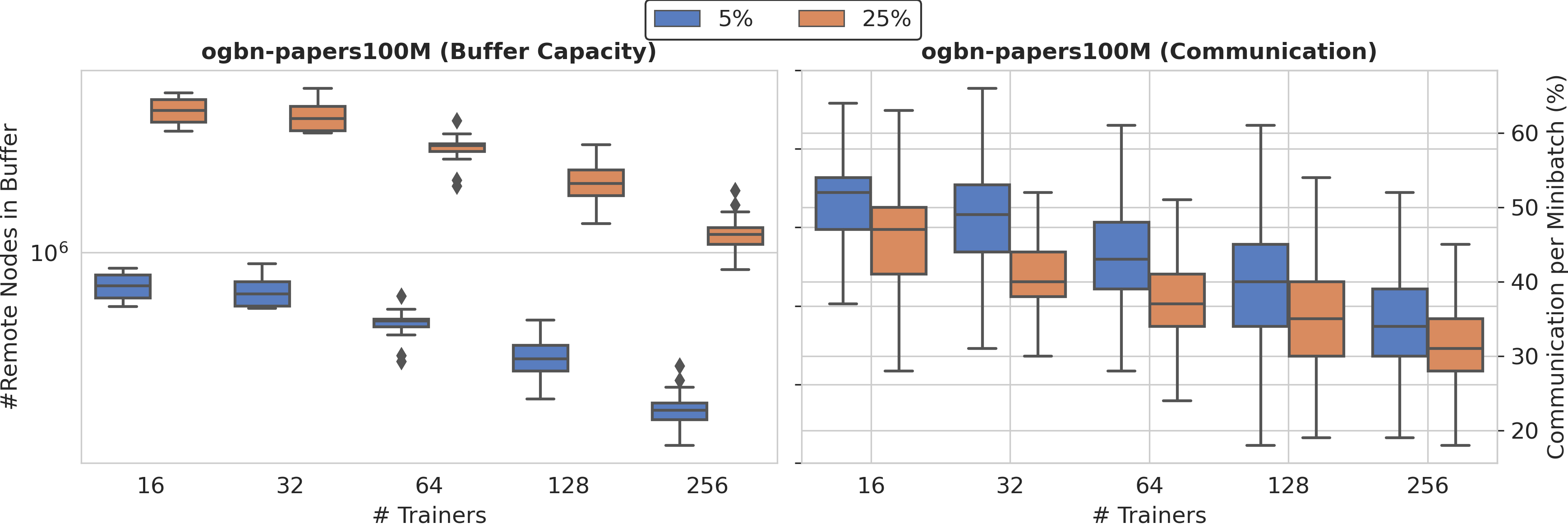}
  \caption{(Left) \#Remote nodes in buffer (5\%\slash 25\%). (Right) 99\% communication volume (\emph{lower} is better).}
 \label{fig:cachesize}
 \end{figure}

We evaluate training performance as mean epoch time across 16--256 trainer PEs (4\slash node), under two persistent buffer sizes: 5\%/25\% (of \emph{remote nodes} relative to the total remote nodes per partition).
For medium inputs (orkut, products, reddit), we use up to 64 trainers (16 nodes), and up to 256 trainers (64 nodes) for larger ones (papers100M, friendster).
As shown in Fig.~\ref{fig:combined_scale}, baseline DistDGL, lacking data persistence, incurs the highest execution times (about 10--50\% higher relative to the rest) due to unchecked communication overheads. DistDGL+fixed reduces communication and improves performance in most cases with fixed prefetching (by 10--30\% relative to baseline DistDGL), but the static policy causes excessive replacements, leading to worse outcomes for some graphs (see \S\ref{ssec:motiv-replacement}): on reddit DistDGL-fixed performs about 35\% worse than the baseline. In comparison, DistDGL+Rudder achieves about 10\% improvement (over DistDGL+fixed) on 64 trainers for reddit, while consistently improving \%-Hits by $\sim$20--50\%. We observe minor performance (and \%-Hits) differences between 5\% and 25\% buffer sizes for large graphs (i.e., papers100M and friendster), but opposite is true for rest of the small-medium graphs (about 30\% better in 25\% buffer sizes for resolving capacity misses) which are more susceptible to communication. Consequently, there can be limited scope in \%-Hits improvement with 5\% buffer if the \#nodes replaced per round across PEs are consistently low, which is the case with the small-medium graphs (e.g., on reddit less than 100 nodes are replaced per round regardless of \#PEs with 5\% buffer). While large graphs on higher \#PEs can exhibit $T_{DDP}$/$T_{COMM}$ overlap (\S\ref{ssec:method-perfmodel}) leading to the best outcome in terms of performance, it may not linearly scale with increasing the \#PEs. A side effect of faster training can result in $T_{DDP}<T_{A\slash C}$ (Remark~\ref{tradeoff-item:strong-scaling},\ref{tradeoff-item:dim-overlap}), i.e., more training steps occuring between successive decisions, lowering the frequency of metric evaluations (e.g., papers100M on 256 PEs, inference is more expensive for LLM than ML Classifier in this case). 
\begin{figure}[!ht]
  \centering
  \includegraphics[width=\linewidth]{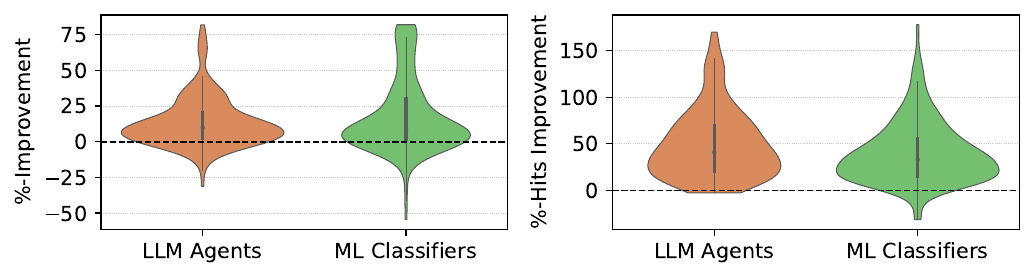}
  \caption{\%-Improvement of Rudder with LLM Agents and ML Classifiers over DistDGL+fixed across all configurations.}
 \label{fig:comparison_spider}
 \end{figure}
Fig.~\ref{fig:comparison_spider} compares the performance spectrum of DistDGL-Rudder considering LLM agents and ML Classifier variants (for varied datasets, buffer sizes and \#trainers) against DistDGL-fixed, depicting median performance improvements in GNN training times of about 10\% and 50\% higher \%-Hits. For the baseline results, we considered a buffer capacity that maximizes persistence, but greater than 2$\times$ performance improvements are possible by tolerating 10--15\% decline in \%-Hits (as discussed in \S\ref{ssec:results-buffer-sizes}). LLM agents have an edge due to no offline training and handling out-of-order distributions (leading to low variability in Fig.~\ref{fig:comparison_spider}); impact on unseen datasets is discussed in \S\ref{ssec:results-finetune}.
\\
\textbf{\textit{Buffer capacity and communication}}
 Fig.~\ref{fig:cachesize} (right) shows the percentage of the remote nodes communicated per minibatch for 5\slash 25\% buffers using Gemma3-4B. 
 Smaller buffers (5\%) incur higher communication (limited capacity), up to 50\% of the sampled nodes; but, \%-communication decreases by 30--40\% with increasing \#trainers. 
\\
\textbf{\textit{Comparison with MassiveGNN}} We compare Rudder with MassiveGNN~\cite{sarkar2024massivegnn}, also built using DistDGL, which initially prefetches high-degree remote nodes prior to training (unlike Rudder which starts with zero elements in persistent buffer), employing heuristics and hyperparameters for periodic replacement. 
 \begin{figure}[!ht]
  \centering
  \includegraphics[width=\linewidth]{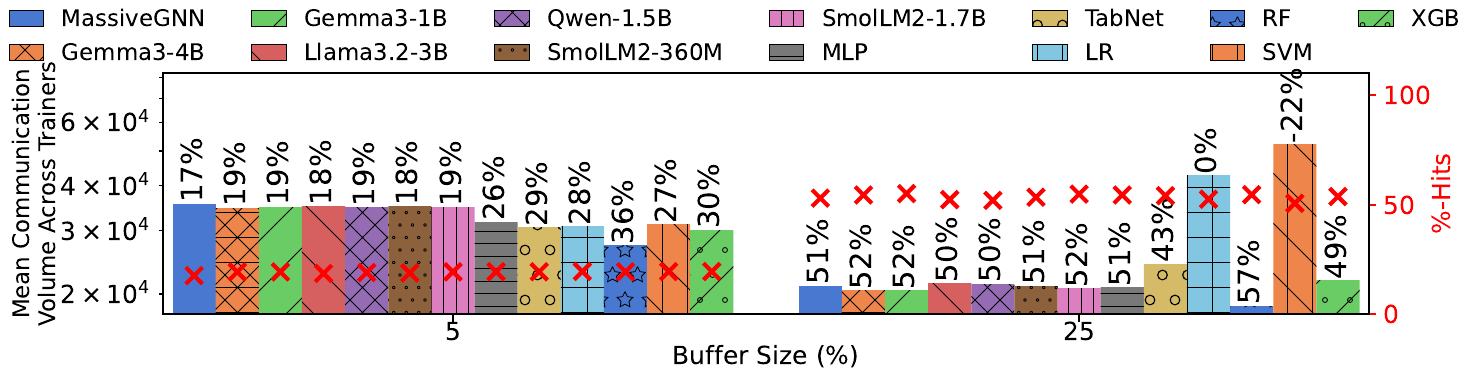}
  \caption{Communication volume (\emph{lower} is better) and \%-Hits for MassiveGNN (fixed replacement interval at 32) and DistDGL+Rudder across buffer sizes (5\%\slash 25\%) on products (64 trainers).}
 \label{fig:comparison}
 \end{figure}
Considering the best reported MassiveGNN hyperparameters (obtained from exhaustive trial-and-error), Rudder demonstrates competitive reduction in the overall communication volume while maintaining high \%-Hits, as shown in Fig.~\ref{fig:comparison}. On products with 64 trainers, Rudder reduces the mean communication by about 19--36\% with 5\% buffers and 43--52\% with 25\% buffers (vs.\ 17\% and 51\% for MassiveGNN) relative to DistDGL (no prefetch). As discussed next in \S\ref{ssec:results-buffer-sizes}, Rudder can reduce communication\slash epoch by greater than 50\% when buffer capacity is limited (however, \%-Hits are higher with larger capacities).\vspace{-0.2cm}



\subsection{Performance and Persistence tradeoffs}\label{ssec:results-buffer-sizes}

\begin{figure}[!ht]
  \centering
      \includegraphics[width=\linewidth]{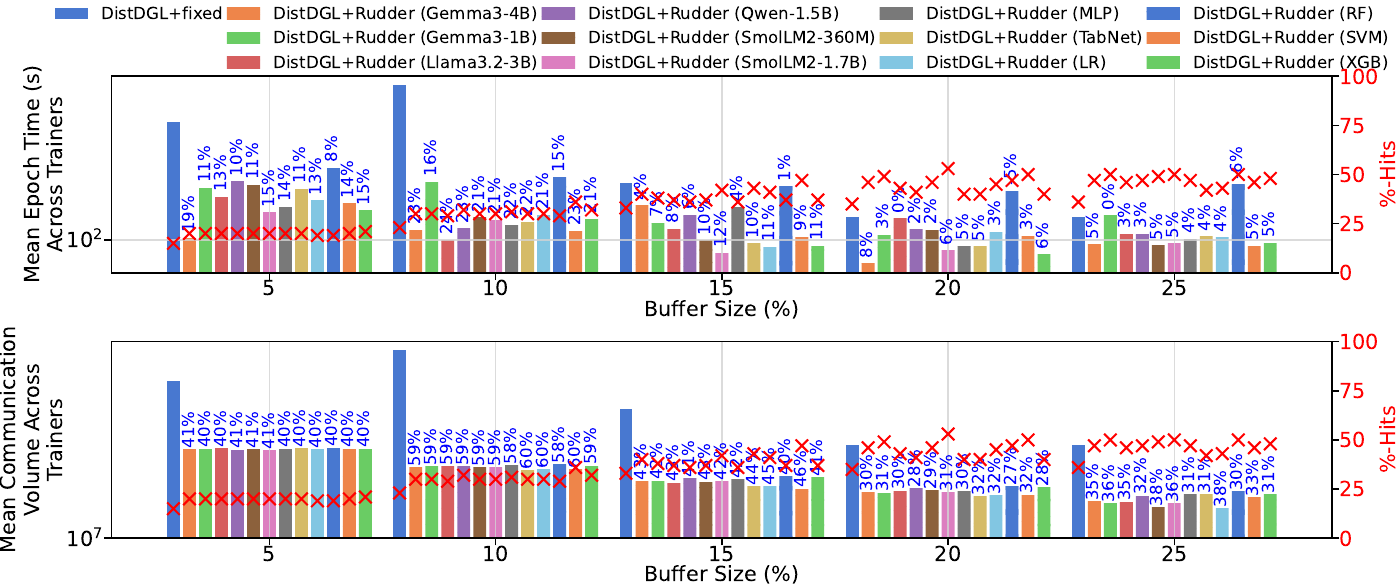}
  \caption{\%-Training time (top) and \%-communication (bottom). Y1: \emph{lower} is better and  Y2 (\%-Hits): \emph{higher} is better, across buffer capacities (5--25\% of remote nodes) in products on 16 trainers. Annotations are \%-improvements relative to DistDGL+fixed.}
  \label{fig:buffersize}
\end{figure}

Fig.~\ref{fig:buffersize} shows the impact of buffer sizes (5--25\%) on mean training time and communication volume in products (16 trainers). DistDGL+fixed, with static prefetching, is slower than Rudder and exhibits highest communication volume, indicating suboptimal data persistence. Gemma3-4B, SmolLM2-1.7B, MLP, and Llama3.2-3B demonstrate performance improvements of 15--19\%, with 5\% buffers (predominant communication). Gemma3-4B exhibits 10--15\% better performance than the rest (without affecting \%-Hits). Overall, we observe about 2--4$\times$ performance improvement potential with 5--15\% buffer capacity (relative to 25\% capacity in Fig.~\ref{fig:combined_scale}), at the expense of overall persistence. 

\subsection{Synchronous vs. Asynchronous}\label{ssec:results-latency}
 \begin{figure}[t]
  \centering
  \includegraphics[width=\linewidth]{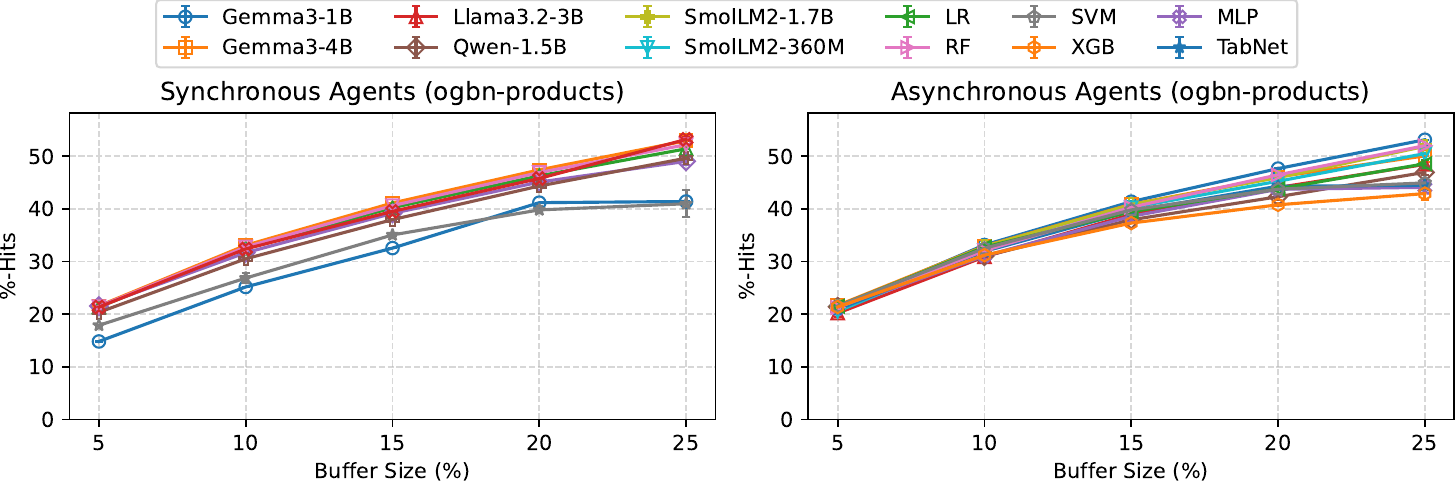}
  \caption{\%-Hits comparison between synchronous (left) and asynchronous (right) mode of Rudder (\emph{higher} is better).}
 \label{fig:sync_vs_async}
\end{figure}
\begin{table}[b]
\caption{Asynchronous vs. Synchronous Evaluations.}
\label{tab:model_performance}
\resizebox{\columnwidth}{!}{%
\begin{tabular}{|ccccc|}
\hline
\multicolumn{1}{|c||}{\multirow{2}{*}{\textbf{Models}}} & \multicolumn{4}{c|}{\textbf{Asynchronous}}                                                                                                                                                                                                                                                                                                                                  \\ \cline{2-5}
\multicolumn{1}{|c||}{}                               & \multicolumn{1}{c|}{\textbf{\begin{tabular}[c]{@{}c@{}}Pass@1 \\ \%-Hits/Acc\end{tabular}}} & \multicolumn{1}{c|}{\textbf{\begin{tabular}[c]{@{}c@{}}Replacement \\ Interval ($r$)\end{tabular}}} & \multicolumn{1}{c|}{\textbf{\begin{tabular}[c]{@{}c@{}}Valid/Invalid \\ Response (\%)\end{tabular}}} & \textbf{\begin{tabular}[c]{@{}c@{}}+ve/-ve \\ Decisions (\%)\end{tabular}} \\ \hline
\multicolumn{1}{|c||}{Gemma3-4B} & \multicolumn{1}{c|}{79} & \multicolumn{1}{c|}{10} & \multicolumn{1}{c|}{100/0}  & 30/70  \\ \hline
\multicolumn{1}{|c||}{Gemma3-1B} & \multicolumn{1}{c|}{16} & \multicolumn{1}{c|}{8} & \multicolumn{1}{c|}{100/0} & 100/0 \\ \hline
\multicolumn{1}{|c||}{Llama3.2-3B} & \multicolumn{1}{c|}{63} & \multicolumn{1}{c|}{6} & \multicolumn{1}{c|}{99/1} & 29/71 \\ \hline
\multicolumn{1}{|c||}{SmolLM2-360M} & \multicolumn{1}{c|}{13} & \multicolumn{1}{c|}{4} & \multicolumn{1}{c|}{87/13} & 35/65 \\ \hline
\multicolumn{1}{|c||}{SmolLM2-1.7B} & \multicolumn{1}{c|}{24} & \multicolumn{1}{c|}{5} & \multicolumn{1}{c|}{92/8} & 70/30 \\ \hline
\multicolumn{1}{|c||}{Qwen-1.5B} & \multicolumn{1}{c|}{38} & \multicolumn{1}{c|}{26} & \multicolumn{1}{c|}{44/56} & 68/32 \\ \hline \hline
\multicolumn{1}{|c||}{MLP} & \multicolumn{1}{c|}{57} & \multicolumn{1}{c|}{1} & \multicolumn{1}{c|}{-} & 12/88 \\ \hline 
\multicolumn{1}{|c||}{TabNet} & \multicolumn{1}{c|}{54} & \multicolumn{1}{c|}{1} & \multicolumn{1}{c|}{-} & 7/93 \\ \hline 
\multicolumn{1}{|c||}{Linear Regression (LR)} & \multicolumn{1}{c|}{52} & \multicolumn{1}{c|}{1} & \multicolumn{1}{c|}{-} & 4/96  \\ \hline
\multicolumn{1}{|c||}{Random Forest (RF)} & \multicolumn{1}{c|}{54} & \multicolumn{1}{c|}{2} & \multicolumn{1}{c|}{-} & 100/0 \\ \hline 
\multicolumn{1}{|c||}{Support Vector Machine (SVM)}  & \multicolumn{1}{c|}{52} & \multicolumn{1}{c|}{2} & \multicolumn{1}{c|}{-} & 4/96 \\ \hline 
\multicolumn{1}{|c||}{XGBoost (XGB)} & \multicolumn{1}{c|}{53} & \multicolumn{1}{c|}{1} & \multicolumn{1}{c|}{-} & 5/95    \\ \hline 
\hline
\multicolumn{5}{|c|}{\textbf{Synchronous}} \\ \hline \hline
\multicolumn{1}{|c||}{Gemma3-4B} & \multicolumn{1}{c|}{83} & \multicolumn{1}{c|}{1} & \multicolumn{1}{c|}{99/1} & 19/81 \\ \hline
\multicolumn{1}{|c||}{Gemma3-1B} & \multicolumn{1}{c|}{3} & \multicolumn{1}{c|}{1} & \multicolumn{1}{c|}{77/23} & 100/0 \\ \hline
\multicolumn{1}{|c||}{Llama3.2-3B} & \multicolumn{1}{c|}{62} & \multicolumn{1}{c|}{1} & \multicolumn{1}{c|}{99/1} & 35/65 \\ \hline
\multicolumn{1}{|c||}{SmolLM2-360M} & \multicolumn{1}{c|}{8} & \multicolumn{1}{c|}{1} & \multicolumn{1}{c|}{87/13} & 35/65 \\ \hline
\multicolumn{1}{|c||}{SmolLM2-1.7B} & \multicolumn{1}{c|}{16} & \multicolumn{1}{c|}{1} & \multicolumn{1}{c|}{91/9} & 75/25 \\ \hline
\multicolumn{1}{|c||}{Qwen-1.5B} & \multicolumn{1}{c|}{42} & \multicolumn{1}{c|}{1} & \multicolumn{1}{c|}{6/94} & 56/44 \\ \hline \hline
\multicolumn{1}{|c||}{MLP} & \multicolumn{1}{c|}{52} & \multicolumn{1}{c|}{1} & \multicolumn{1}{c|}{-} & 10/90 \\ \hline
\multicolumn{1}{|c||}{TabNet} & \multicolumn{1}{c|}{47} & \multicolumn{1}{c|}{1} & \multicolumn{1}{c|}{-} & 4/96 \\ \hline
\multicolumn{1}{|c||}{Linear Regression (LR)} & \multicolumn{1}{c|}{52} & \multicolumn{1}{c|}{1} & \multicolumn{1}{c|}{-} & 9/91 \\ \hline
\multicolumn{1}{|c||}{Random Forest (RF)} & \multicolumn{1}{c|}{55} & \multicolumn{1}{c|}{1} & \multicolumn{1}{c|}{-} & 100/0 \\ \hline 
\multicolumn{1}{|c||}{Support Vector Machine (SVM)} & \multicolumn{1}{c|}{52} & \multicolumn{1}{c|}{1} & \multicolumn{1}{c|}{-} & 9/91 \\ \hline 
\multicolumn{1}{|c||}{XGBoost (XGB)} & \multicolumn{1}{c|}{50} & \multicolumn{1}{c|}{1} & \multicolumn{1}{c|}{-} &  6/94    \\ \hline 
\end{tabular}%
}
\end{table}

\begin{figure*}[!t]
  \centering
  \begin{minipage}[t]{0.30\linewidth}
    \centering
    \includegraphics[width=0.48\linewidth]{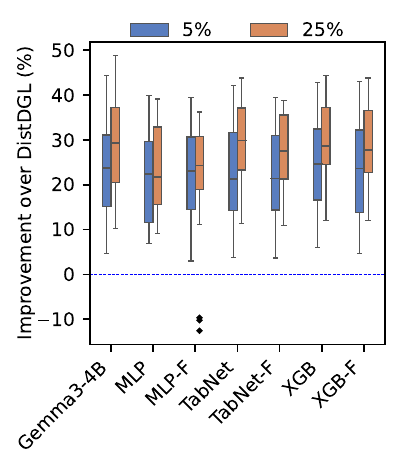}%
    \includegraphics[width=0.48\linewidth]{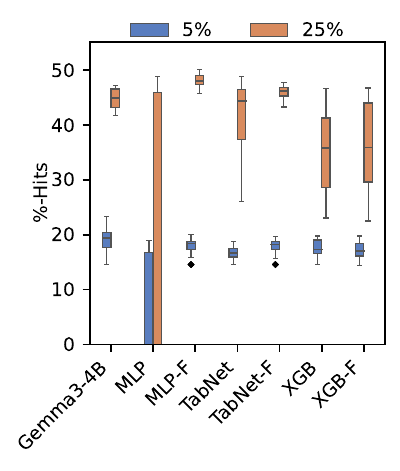}
    \captionof{figure}{Yelp (8--64 trainers)}
    \label{fig:yelp}
  \end{minipage}%
  \begin{minipage}[t]{0.30\linewidth}
    \centering
    \includegraphics[width=0.48\linewidth]{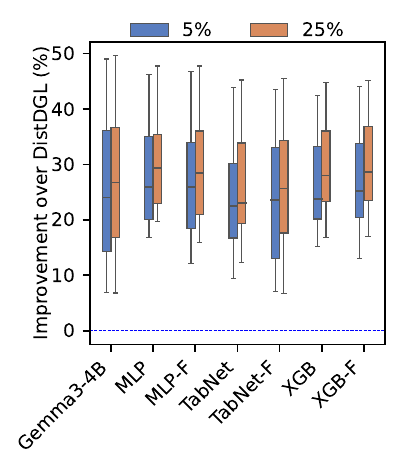}%
    \includegraphics[width=0.48\linewidth]{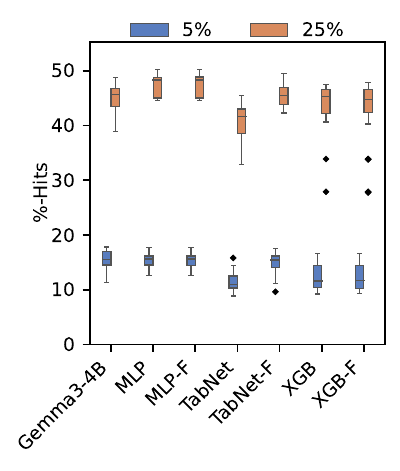}
    \captionof{figure}{arxiv (8--32 trainers)}
    \label{fig:arxiv}
  \end{minipage}
  \begin{minipage}[t]{0.35\linewidth}
    \vspace*{-2.5cm} 
    \centering
    \scriptsize
    \renewcommand{\arraystretch}{0.9}
    \captionof{table}{Pass@1 \%-Hits (\textit{higher} is better) of Gemma3-4B and Acc of ML Classifiers (includes finetuned variants) across 1000-2000 batchsizes.}
    \label{tab:unseen}
    \begin{tabular}{|c||c|c|c|c|}
      \hline
      \textbf{Datasets} & \textbf{Gemma3-4B} & \textbf{MLP/F} & \textbf{TabNet/F} & \textbf{XGB/F} \\
      \hline\hline
      yelp       & 48 (-8/9) & 51 (-5/6) & 50 (-5/5) & 50 (5/5) \\ \hline
      ogbn-arxiv & 52 (-17/19) & 52 (-4/5) & 49 (-4/4) & 50  (-2/2) \\ \hline
    \end{tabular}
  \end{minipage}
\end{figure*}
Synchronous deployment of agents\slash classifiers provides a consistent view of the system but stalls trainers during request processing (see \S\ref{ssec:method-syncasync}), increasing $T_{DDP}$ by up to $25\times$ (e.g., Qwen1.5B). Such overheads outweigh the modest \%-Hits gains (<5\% across buffer sizes), except Gemma3-1B, which shows $\sim$10\% improvement (Fig.~\ref{fig:sync_vs_async}, products, 16 trainers). MLP’s dense mapping remains stable when applied every minibatch in synchronous mode, but degrades by $\sim$10\% in asynchronous version with slightly fewer decisions. TabNet’s sparse gating mechanism is ineffectual in synchronous mode (discards useful features often), which improves slightly in asynchronous mode by $\sim$8\%, where fewer requests reduce such errors.
Table~\ref{tab:model_performance} presents model performance under asynchronous\slash synchronous modes (see \S\ref{ssec:method-eval}). For classifiers, we report accuracy instead of Pass@1, since predictions are supervised against labeled traces rather than evaluated through the outcome alignment. Gemma3-4B achieves the highest Pass@1 \%-Hits (about 80\%) and Llama3.2, the second highest (conservatively suggesting replacements 19--30\% of the instances).
Our prompt adheres to JSON format, and the three Llama-based models (Gemma3-4B/1B and Llama3.2-3B) exhibited compliance, resulting in near 100\% valid responses. However, Gemma3-1B's responses reveal poor reasoning--as \%-Hits rise, it repeatedly infers a decline in performance from context, initiating replacements aggressively. It mimics DistDGL+fixed in synchronous mode, resulting in the least Pass@1. On large graphs with dense neighborhoods (i.e., likely to sample higher unique nodes), such aggressiveness can briefly raise \%-Hits (e.g., first 50 epochs), a phenomenon we term ``replacement bias.''
Llama3.2-3B achieves high accuracy with least latency (i.e., $T_{A\slash C} < T_{DDP}$), making it ideal for latency-sensitive decisions. In contrast, Qwen-1.5B shows longer \emph{replacement intervals} (26, see \S\ref{ssec:method-dist-alg}) and the least valid response rate (44\% async, 6\% sync), lacking in both performance and accuracy. SmolLMs, though fastest, perform poorly in Pass@1, underscoring that latency alone is insufficient; reasoning quality is equally critical for maintaining balance. Replacement intervals of ML classifiers are relatively low, suggesting higher decision frequency (due to faster inference in most cases, synchronous and asynchronous performances and accuracies are also comparable), which can adversely affect the overall communication volume, as discussed in \S\ref{ssec:results-llm-reasoning}. The low accuracy for classifiers is also indicative of the challenges in curating ground truth labels for assessment.
In Table~\ref{tab:async_performance}, we report Pass@1 \%-Hits for all models\slash datasets in asynchronous mode. Gemma3-4B consistently shows the highest score. We also show 95\% confidence intervals (CI) per run, computed via chi-square distribution, to capture variability: wider CIs indicate higher fluctuation, narrower CIs reflect stable performance.
\begin{table}[h]
\caption{Pass@1 \%-Hits/Accuracy (\emph{higher} is better, +95\% CI) reflecting per-run variability for Rudder (async).}
\label{tab:async_performance}
\resizebox{\columnwidth}{!}{%
\begin{tabular}{|c||c|c|c||c|c|}
\hline
\textbf{Models} & \textbf{products} & \textbf{reddit} & \textbf{papers} & \textbf{orkut} & \textbf{friendster} \\ \hline \hline
Gemma3-4B    & 76 (-9/11)       & 75 (-12/15)    & 74 (-8/9)      & 82 (-10/12)   & 77 (-7/8)          \\ \hline
Gemma3-1B    & 20 (-16/19)      & 19 (-12/15)    & 13 (-11/12)    & 9 (-13/15)    & 82 (-7/8)          \\ \hline
Llama3.2-3B  & 62 (-11/13)      & 63 (-10/13)    & 64 (-10/11)    & 64 (-11/14)   & 58 (-11/13)        \\ \hline
SmolLM2-360M & 14 (-20/24)      & 13 (-14/17)    & 13 (-18/20)    & 9 (-14/17)    & 52 (-11/13)        \\ \hline
SmolLM2-1.7B & 25 (-13/16)      & 24 (-13/16)    & 25 (-13/14)    & 20 (-14/17)   & 68 (-12/13)        \\ \hline
Qwen-1.5B    & 36 (-21/26)      & 34 (-20/24)    & 38 (-15/17)    & 36 (-17/20)   & 35 (-22/24)        \\ \hline \hline
MLP    & 62 (-8/10)      & 62 (-10/12)    & 53 (-12/13)    & 50 (-3/3)   & 47 (-11/13)        \\ \hline 
TabNet    & 58 (-8/9)      & 59 (-11/14)    & 51 (-10/11)    & 62 (-8/10)   & 77 (-10/11)        \\ \hline
LR    &   57 (-10/12)    &  58 (-11/14)   & 59 (-11/13)    & 57 (-5/7)   &   87 (-8/9)    \\ \hline
RF    &   62 (-8/10)    &  62 (-10/12)   &  55 (-4/4)   &  52 (-2/3)  &    91 (-4/5)     \\ \hline
SVM    &  57 (-11/14)     &  56 (-9/11)   &  56 (-14/15)   &  57 (-8/10)  &     85 (-8/9)    \\ \hline
XGB    &   58 (-8/9)    &  61 (-10/12)   &  52 (-8/9)   &  65 (-9/11)  &    79 (-10/11)     \\ \hline
\end{tabular}%
}
\end{table}

\textbf{\textit{Replacement bias}}
As partitions shrink, a trainer may only encounter a single minibatch\slash epoch, reducing potential decision instances. Large inputs like friendster may exhibit startup issues from higher memory consumption, requiring more partitions as a baseline. But, models may still report high Pass@1 \%-Hits due to \emph{replacement bias} (e.g., Gemma3-1B\slash RF reports Pass@1 \%-Hits of 82\%\slash 91\% for friendster on 50 epochs). With a single minibatch\slash epoch (training set limited to top-5000 communities), sampling from a large graph such as friendster often yields unique nodes, so nearly every replacement yields \emph{instant gratification} in \%-Hits, at least during the earlier epochs (delaying buffer stabilization). Expanding to 100 epochs, Gemma3-1B \%-Hits drops to 51\%, while Gemma3-4B, maintains at 70\%.

\subsection{Performance on Unseen Datasets}\label{ssec:results-finetune}
We evaluate performance on unseen (excluded in offline training of the ML Classifiers presented in Eqn~\ref{eqn:cost_icl}) Yelp and ogbn-arxiv dataset with Gemma3-4B, pretrained MLP, TabNet, XGB and their finetuned variants across batch sizes (500, 1000, 2000), \#trainers, and buffer sizes (5\%, 25\%) (Fig.~\ref{fig:yelp}, \ref{fig:arxiv}). Finetuning is triggered at every 5\slash 25\slash 50 minibatches (selected empirically). Smaller batches in fewer trainers produce more \#minibatches\slash trainer, lowering per-minibatch communication (fewer remote nodes sampled). This distribution shift (Remark~\ref{tradeoff-item:dist-shift}) from MLP’s training setting (batchsize 2000) causes lower \%-Hits (often zero, represented by the wide range in Fig.~\ref{fig:yelp}) and inflates the training time. Even at zero \%-Hits, MLP outperforms DistDGL, not from reduced communication (empty buffer), but from overlapping future minibatch preparation with GNN training. Periodic finetuning improves \%-Hits, but adds computation overhead (8-trainers/500-batch runs, Fig.~\ref{fig:yelp}, \ref{fig:arxiv}). Gemma3-4B maintains high \%-Hits with lower epoch times (Corollary~\ref{lemma:decision-quality}). With more trainers and small batches (e.g., 128 trainers, 500\slash 1000 batchsize), due to fewer and smaller minibatches (Remark~\ref{tradeoff-item:strong-scaling}) overlap opportunities are limited (Remark~\ref{tradeoff-item:dim-overlap}).

\subsection{Replacement Trajectories of LLM vs. ML}\label{ssec:results-llm-reasoning}
\begin{figure}[!ht]
  \centering
    \includegraphics[width=0.49\linewidth]{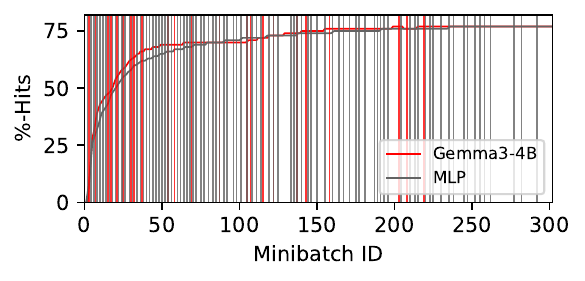}
     \includegraphics[width=0.49\linewidth]{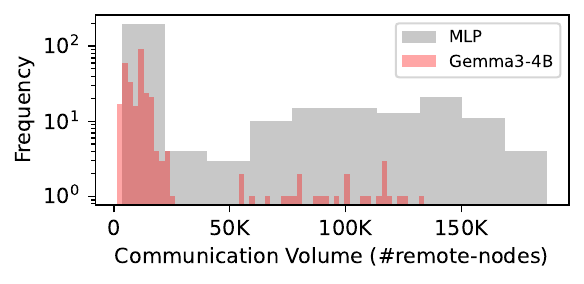}
  \caption{Trajectories of \%-Hits (left) and communication volume (right) of papers100M on 64 nodes with LLM and MLP.}
   \label{fig:llm_reasoning}
\end{figure}

To assess the integrity of LLM's replacement strategy, we compare the temporal trajectories of Gemma3-4B and MLP on a single trainer (with comparable steady-state \%-Hits) in Fig.~\ref{fig:llm_reasoning}. As expected, in both cases, \%-Hits increases rapidly during the initial minibatches, eventually converging to a steady state. However, LLM decides on replacement selectively when the evolving trajectory indicates that the current state is suboptimal (possibly due to suboptimal\slash stagnating \%-Hits and rising communication), as evidenced from relatively infrequent interventions during the later minibatches. 
In contrast, the MLP (and other ML classifiers in similar workloads) is trained as a pointwise classifier, and invokes replacement more frequently throughout the GNN training (many more gray vertical lines than red in Fig.~\ref{fig:llm_reasoning}), lacking an implicit notion of long-horizon tradeoff, and continuing to replace with diminishing returns. A side effect of frequent replacement (\S\ref{ssec:motiv-replacement}) in MLP can be orders of magnitude increase in overall communication relative to LLM agents (Fig.~\ref{fig:llm_reasoning}).\vspace{-0.2cm}

\subsection{Mixture of Experts as LLM agents}\label{ssec:results-moe}
\begin{figure}[h]
  \centering
    \includegraphics[width=\linewidth]{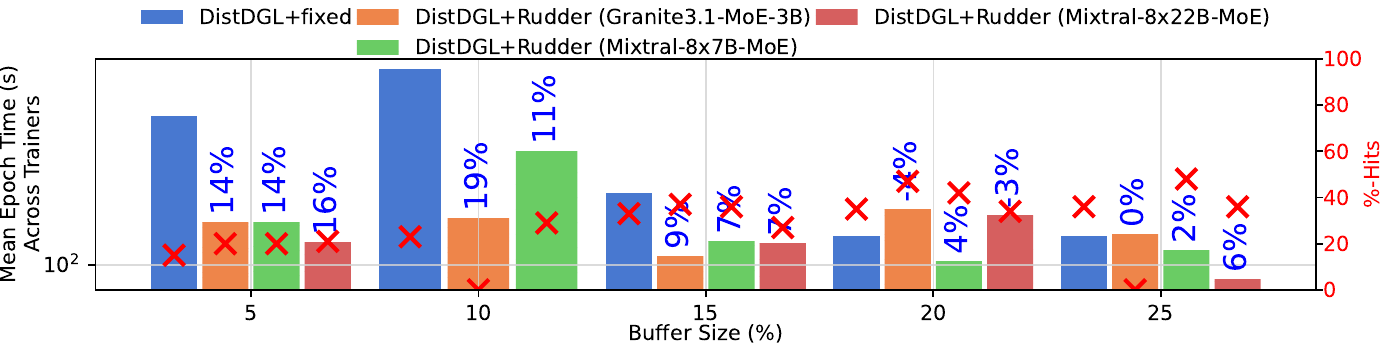}
  \caption{Training times (\emph{lower} is better) for MoEs across 5--25\% buffer sizes on 16 trainers. Note: Mixtral-8x22B stalled in 10\%.}
   \label{fig:buffersize:moetraining}
\end{figure}

We evaluated Mixture of Experts (MoE) models in Rudder, including Granite3.1~\cite{granite2024granite}, Mixtral-8x7B, and Mixtral-8x22B~\cite{jiang2024mixtral} on NERSC Perlmutter nodes with 80GB NVIDIA A100 GPUs on products. Despite exploiting sparsity to reduce resource requirements, MoEs offer limited benefit and do not outperform lightweight models like Gemma3-4B (Pass@1 \%-Hits <60\% in Table~\ref{tab:moe_performance}). 
MoE models improve only at smaller buffer sizes (5--15\%), with up to 20\% gains over DistDGL+fixed, but little benefit beyond buffer size 20\% (Fig.~\ref{fig:buffersize:moetraining}). Mixtral-8x22B (the smallest quantized version is 52GB), often stalled and froze at 10\% buffer on 80GB A100s GPUs. Despite being the largest model tested, it performed worst among MoE agents as low-bit quantization significantly degrades reasoning in large models~\cite{li2025quantization, liu2023emergent}. Thus, larger models do not guarantee better results in performance-sensitive workloads.
\begin{table}[h]
\caption{Pass@1 \%-Hits (\emph{higher} is better) of different-sized MoE models using products on 16 trainers.}
\label{tab:moe_performance}
\centering
\scriptsize
\begin{tabular}{lcccc}
\toprule
\textbf{MoE LLMs} 
  & \shortstack{\textbf{Pass@1} \\ \textbf{\%-Hits}} 
  & \shortstack{\textbf{Replacement} \\ \textbf{Interval ($r$)}}
  & \shortstack{\textbf{Valid/Invalid} \\ \textbf{Response (\%)}}
  & \shortstack{\textbf{(+ve/-ve)} \\ \textbf{Decisions (\%)}}\\
\midrule
Granite3.1-3B       
    &  44  & 21  & 99/1 & 61/39  \\

Mixtral-8x7B       
    & 50   & 21   & 94/6  & 56/44 \\

Mixtral-8x22B    
    &  56  & 42 & 100/0 & 86/14 \\
\bottomrule
\end{tabular}
\end{table}

\section{Conclusion}
Rudder exposes flexible trade-offs (on top of an optimized asynchronous baseline) to choose between the desired level of data persistence and communication optimization relative to state-of-the-art fixed and heuristics driven prefetching policies, leading to 50\% communication reduction and 90\% improvement in minibatch-driven GNN training performance. The core problem that Rudder solves is optimizing prefetch parameters for a massive space that is ill-defined because gathering training data is experimentally\slash computationally expensive, making it ideal for LLMs. Our work has a transferable insight because LLMs as agents are good at approximate search of an extremely large space, without not only derivatives but a formal goodness function (In-Context Learning). We demonstrate that even small LLMs (manageable memory footprint) can perform well on this otherwise laborious\slash expensive problem. Unlike ML classifiers, LLMs can perform well on  out-of-distribution problems.

\section{Acknowledgement}
This research is supported by the National Science Foundation (NSF) under Award 2243775 and the U.S. Department of Energy (DOE) through the Office of Advanced Scientific Computing Research's ``Orchestration for Distributed \& Data-Intensive Scientific Exploration'' and ``Orchestrated Platform for Autonomous Laboratories'' for Foundational AI Models for Optimizing and Understanding Biological Systems (OPAL-FAMOUS) projects. Pacific Northwest National Laboratory is operated by Battelle for the DOE under Contract DE-AC05-76RL01830. This research used resources of the National Energy Research Scientific Computing Center (NERSC), a Department of Energy User Facility. We thank Amazon Web Services (AWS) for providing cloud computing credits in support of this research.
\bibliographystyle{ACM-Reference-Format}
\bibliography{paper/references}

\end{document}